\documentclass{article}

\usepackage{arxiv}

\usepackage[utf8]{inputenc} 
\usepackage[T1]{fontenc}    
\usepackage{hyperref}       
\usepackage{url}            
\usepackage{booktabs}       
\usepackage{amsmath}  
\usepackage{amsfonts}       
\usepackage{nicefrac}       
\usepackage{microtype}      
\usepackage{makecell}           
\usepackage{lipsum}
\usepackage{graphicx}
\graphicspath{ {./images/} }
\usepackage{multirow}           
\usepackage{tabularx}           

\usepackage{lipsum}             
\usepackage{xspace}             
\usepackage{pdflscape}          

\usepackage{setspace}           
\usepackage{indentfirst}        
\title{DeepWheel: Generating a 3D Synthetic Wheel Dataset for Design and Performance Evaluation}

\author{Soyoung Yoo\\
	Cho Chun Shik Graduate School of Mobility\\
	KAIST\\
	Daejeon, 34051\\
	\texttt{soyngyoo@kaist.ac.kr} \\
	\And
	Namwoo Kang\\
	Cho Chun Shik Graduate School of Mobility\\
	KAIST\\
        Narnia Labs\\
	Daejeon, 34051\\
	\texttt{nwkang@kaist.ac.kr} \\
      }

\begin{document}
\maketitle

\begin{abstract}
Data-driven design is emerging as a powerful strategy to accelerate engineering innovation. However, its application to vehicle wheel design remains limited due to the lack of large-scale, high-quality datasets that include 3D geometry and physical performance metrics. To address this gap, this study proposes a synthetic design-performance dataset generation framework using generative AI. The proposed framework first generates 2D rendered images using Stable Diffusion, and then reconstructs the 3D geometry through 2.5D depth estimation. Structural simulations are subsequently performed to extract engineering performance data. To further expand the design and performance space, topology optimization is applied, enabling the generation of a more diverse set of wheel designs. The final dataset, named DeepWheel, consists of over 6,000 photo-realistic images and 900 structurally analyzed 3D models. This multi-modal dataset serves as a valuable resource for surrogate model training, data-driven inverse design, and design space exploration. The proposed methodology is also applicable to other complex design domains. The dataset is released under the Creative Commons Attribution-NonCommercial 4.0 International(CC BY-NC 4.0) and is available on the \href{https://www.smartdesignlab.org/datasets}{dataset page}
\end{abstract}

\section{Introduction}

\subsection{Research Background}
Automotive wheels are key components that require careful consideration of engineering performance from the initial design stage. They are directly linked to various vehicle performance factors such as driving performance, fuel efficiency, and safety. Generally, the wheel design starts with a 2D sketch by a designer, gradually adding details and evaluating performance through simulations such as 3D CAD modeling, finite element analysis (FEA), and computational fluid dynamics (CFD) before going through an iterative modification process to determine the final shape. At this point, the simulation results have a decisive influence on the design selection as performance indicators, and a close feedback loop between design and analysis is essential.

However, in the design process, aesthetic preferences and brand identity are prioritized in the early stages \cite{damen2024exploring,feldinger2017automotive}. As a result, large-scale iterative modifications are unavoidable if a design that does not satisfy the expected performance is identified in the analysis stage. This is a major cause of delays in the development schedule and increased costs. To solve these problems, the need for a surrogate model that can quickly predict performance at the initial design stage is emerging. Surrogate models can quickly evaluate various design alternatives without complex simulations, minimizing the number of design iterations and enabling efficient design decisions. \cite{yoo2021,song2023Surrogate} 

Recently, instead of traditional parametric models, research on deep learning-based surrogate models that use the entire 3D shape as input is being actively conducted. This allows for accurate prediction of physical performance even for complex shapes using a non-parametric method, and enables performance prediction for various wheel shapes. \cite{shin2023wheel,shin2024uda}. In particular, studies using graph neural networks (GNNs) show high accuracy in predicting engineering characteristics using 3D meshes as input. \cite{kim2025physics, park2024bmo}.

The performance of a deep learning-based surrogate model is greatly affected by the quality and variety of data used for training. However, in the automotive industry, datasets containing high-quality 3D geometry and analysis data are primarily private or have limited access. Therefore, a large dataset containing high-quality and diverse 3D geometry models and analysis (FEA/CFD) results is required to develop a reliable surrogate model. \cite{song2024multi, rad2024datasets, elrefaie2025drivaernet++}

In particular, diversity in shape is significant for 3D shape-based deep learning tasks. \cite{li2022predictive, elrefaie2025drivaernet++} It is difficult to expect sufficient generalization performance with only variations of the same shape, and performance prediction based on changes in analysis conditions is also limited. In other words, a variety of shapes must be secured to enable mapping with various analysis results, and this enables the construction of a more robust surrogate model. \cite{rad2024datasets}

Recent advances in generative foundation models, such as Stable Diffusion and DALL·E [11,12], have made it possible to generate high-quality 2D renderings rapidly. In addition, technologies that can predict 3D shapes from real-world images are also advancing. \cite{Get3d2022,Magic3d2023,PointE2022,Zero-1-to-3_2023,One-2-3-45_2023,Triposr2024,Sf3d2024}. This technology can be used to automatically build large-scale design datasets without human intervention, and it can be an innovative approach to solving the existing data shortage problem.

\subsection{Research Objective}
This study proposes a fully automated framework for generating wheel design data using generative AI. The framework aims to create a comprehensive dataset that connects realistic design generation from based images, 3D shape reconstruction, and structural analysis.

\subsection{Research Contributions}
This study proposes a new 3D design data generation methodology that utilizes foundation models in the field of engineering design to support data-driven design and makes the following contributions.

\begin{enumerate}
    \item \textbf{Generative AI-based 3D synthetic wheel generation framework} \\
    This study proposes a generative AI-based dataset generation framework to address the lack of existing 3D wheel design data. The proposed framework consistently and automatically generates and links 2D mask images, rendered images, and 3D wheel shapes. This approach enables more efficient construction of large-scale datasets and can be widely applied to various data-driven design domains.    
    
    \item \textbf{Increasing structural and geometric diversity to explore a wider design space.} \\ 
    By applying topology optimization to reference shapes to generate a variety of structural shape candidates and converting them into photo-realistic rendered forms using a pre-trained Stable Diffusion-based image-to-image translation technique, structural and volumetric diversity previously challenging to achieve with conventional approaches can be realized. This process enables the creation of a broad design space that allows for rapid exploration of various design candidates.  
        
    \item \textbf{Foundation-based depth map prediction and design space embedding} \\
    By training a foundation  model that demonstrates high generalization performance with only a small amount of training data, depth geometry is accurately predicted from rendered wheel designs. The resulting depth maps are embedded into the latent space and are effectively utilized to represent the overall design space.

    \item \textbf{Large-scale multi-modal wheel design–performance dataset for data-driven design} \\
    A large-scale multi-modal wheel design–performance dataset is presented, integrating various types of design data to support data-driven design. The dataset includes a total of 6,249 photo-realistic synthetic rendering images. Additionally, it provides 904 3D mesh models that have successfully undergone finite element analysis (FEA), along with corresponding structural analysis data. For more information and download, please visit the \href{https://www.smartdesignlab.org/datasets}{dataset page}.
\end{enumerate}

\section{Related works}

\subsection{3D Dataset for Engineering Design}

\begin{figure}[!h]
    \centering
    \includegraphics[width=1\linewidth]{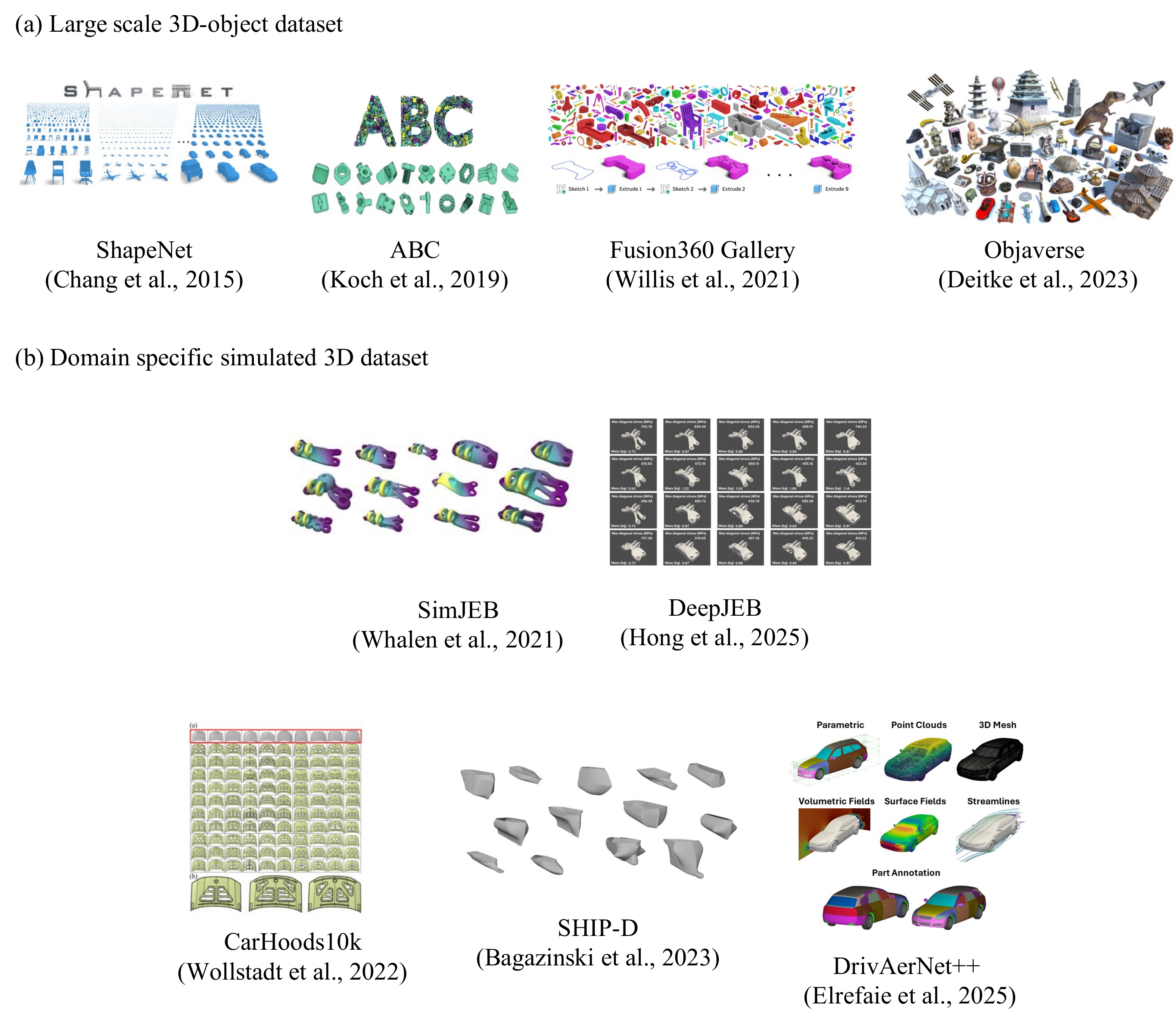}
    \caption{Overview of public 3D object and engineering datasets}
    \label{fig:enter-label}
\end{figure}

3D model datasets are essential for engineering optimization techniques and machine learning model training research. These datasets provide design process and performance evaluation information, effectively supporting design automation and optimization. However, as design criteria change over time or new requirements are added, data consistency becomes difficult to maintain, and collecting additional data is also costly and time-consuming.
The ABC dataset consists of more than one million large-scale CAD models. It is widely used in research in geometric deep learning, such as face normal estimation and surface reconstruction. Objaverse is a large 3D object dataset with various annotations that support machine learning research in various fields, such as engineering design, robotics, and virtual environments. In particular, it has become an important benchmark in research that combines object recognition, and generative design \cite{deitke2023objaverse}. ShapeNet \cite{chang2015shapenet} is a large-scale 3D model repository with various CAD models. The main sub-dataset, ShapeNetCore, provides 51,000 refined 3D models in 55 categories. In comparison, ShapeNetSem provides 12,000 models with additional metadata such as volume and weight, which are widely used in engineering performance optimization and deep learning research \cite{rosset2023,song2023Surrogate}.

A 3D CAD dataset for a typical large-scale category is the Fusion 360 Gallery \cite{willis2021fusion}. This dataset is useful for analyzing repetitive patterns and temporal changes in actual design processes, including 8,625 CAD models and design sequences by human designers. It is especially suitable for developing generative design tools and researching design automation.
There are also specialized datasets for specific engineering fields. The SHIP-D dataset \cite{bagazinski2023ship} contains various hull shapes and hydrodynamic data for ship design optimization and is used to develop machine learning-based performance prediction and design optimization models. The CarHoods10k dataset \cite{wollstadt2022carhoods10k} is specialized for the design of car hoods and provides explicit design parameters and triangular mesh data along with structural dynamics simulation data, making it suitable for research on machine learning-based design optimization models. The SimJEB \cite{whalen2021simjeb} and DeepJEB \cite{hong2025deepjeb} datasets provide actual design data, simulation data, and rich synthetic data, respectively, and are useful for research on optimizing jet engine bracket design. DrivAerNet++ \cite{elrefaie2025drivaernet++} The dataset provides high-fidelity CFD simulation data for analyzing the aerodynamic performance of various geometry types and automotive designs, making it an important benchmark dataset for the development of automotive design optimization and performance prediction models. 

These datasets promote design optimization and automation research in each field and are essential for building an efficient data-driven design process. This study built and provided a multi-modal dataset for data-driven wheel design. 

\subsection{Generative Model for Automotive Wheel Design}

Previous studies on AI-based wheel design have mainly focused on creating 2D binary images using various generative models such as GAN, VAE, and Diffusion Models or automatically generating high-quality rendering images. 

For example, topology optimization results can be trained on GAN to obtain a new wheel structure, which has the advantage of deriving a design with an optimized structure in real-time \cite{oh2019}.
In addition, attempts have been made to maximize the morphological spectrum by exploring the results of optimization of various structures through reinforcement learning \cite{jang2022}. However, since most of these approaches remain at the level of 2D binary images, further work is required to extend them to detailed design proposals. 
On the other hand, approaches that use rendered 2D images are being explored. For example, a process is being proposed that combines machine learning and human aesthetic assessments to explore visual design spaces or that uses the Stable Diffusion model to quickly generate high-quality wheel renderings with only text and image input to inspire designers, as described in \cite{jeon2024} and \cite{wang2024}. 

This approach supports designers in generating initial concepts and visualizing them in various ways; however, an additional step is required to evaluate whether the generated images meet engineering requirements. To achieve this, it is necessary to go beyond the 2D image level, reconstruct 3D geometries, and evaluate their engineering performance. There are ongoing efforts to integrate CAD/CAE systems with deep learning techniques to automate the generation and evaluation of 3D wheel designs. 

Yoo et al. (2021)\cite{yoo2021} proposed a framework that derives a 3D wheel shape from a 2D topology optimization result, converts it into a CAD model, and performs structural analysis. This approach is significant because it enables the generation of diverse design alternatives, extends them into the 3D domain, and verifies their feasibility through automated performance evaluation. On the other hand, in \cite{li2024}, Stable Diffusion was used to automatically generate various 2D wheel mask images and 3D model them in an extrude method in a CAD system, demonstrating the possibility of creating a variety of wheel designs with only a small amount of training data. This approach makes it easy to try out AI-based designs even in situations where large-scale data is challenging to obtain. However, since both studies have relatively simple 3D shapes, there are still limitations in reproducing the details of complex three-dimensional spoke shapes, and further improvement is needed.

\subsection{3D Reconstruction from Image}
Recent advances in computer vision have enabled various approaches to generate 3D models from text and images. The Point-E\cite{PointE2022} study proposed a method for quickly generating 3D point clouds based on the Diffusion model. Magic3D\cite{Magic3d2023} introduced a coarse-to-fine optimization framework to efficiently generate high-resolution texture meshes using sparse 3D neural representations and differentiable rendering techniques. However, the existing approach still has problems, such as a lack of geometric details, limited mesh topology, and texture support.
To address these limitations, GET3D\cite{Get3d2022} was proposed, which successfully generates high-quality texture meshes using only 2D image datasets by combining the advantages of the multi-view Diffusion model and 2D GAN. 
Similarly, Zero-1-to-3\cite{Zero-1-to-3_2023} used the view-conditional diffusion model to generate images from various viewpoints from a single image and finally performed 3D reconstruction based on this. Subsequently, the One-2-3-45\cite{One-2-3-45_2023} study achieved accurate 3D mesh generation in just 45 seconds from a single image. Recent studies have also shown that 3D model generation technology is developing in a direction that significantly improves speed and quality. 
InstantMesh\cite{Instantmesh2024}, TripoSR\cite{Triposr2024}, and SF3D\cite{Sf3d2024} have introduced a feed-forward approach based on neural networks to enable fast and accurate mesh generation. 
In particular, MeshFormer\cite{Meshformer_2025} has realized more sophisticated 3D structure generation by utilizing 3D structures and normal maps through a new architecture that combines Transformer and 3D convolution. One of the biggest challenges of image-based 3D reconstruction is to reconstruct fine geometric details from high-resolution images while maintaining structural consistency. 

Zhang et al. (2024)\cite{M3D2024} demonstrated higher 3D reconstruction performance using a depth map as a guide. Kim et al. (2024)\cite{Kim_mass_2024}'s research showed that innovative designs suitable for mass production can be derived using depth images. 
Estimating depth map based on single-view images is equivalent to estimating depth from a monocular camera, and is essentially an ill-posed problem. Therefore, it has been difficult to collect large-scale ground truth depth data, although it heavily relies on large-scale and diverse training datasets. 
To solve this problem, MiDaS\cite{MiDaS2020} has developed a technique that allows training using a mix of independent datasets with different characteristics and biases, and has demonstrated that training using a mix of various data through zero-shot cross-dataset transfer significantly improves the performance of monocular camera-based depth prediction. Recently, Depth Anything\cite{DepthAnything2024} proposed a foundation model that performs well in images of any condition. 
Instead of adding new technical modules, the focus was on reducing the generalization error by automatically annotating a large-scale unlabeled image dataset of approximately 62 million images to dramatically expand the scope of the dataset. This has set a new standard for stable depth prediction even in high-resolution images. In addition, the Marigold\cite{Marigold2024} methodology proposed a new monocular depth prediction method by utilizing the powerful prior knowledge of Stable Diffusion, which has been successfully used recently. This demonstrated remarkable depth prediction performance and suggested various application possibilities.

This study incorporated a depth map-based 3D reconstruction method into the data generation framework. The proposed method enabled fast 3D reconstruction while preserving high-resolution geometric details of the wheel structure. Its performance was validated by comparison with existing image-to-3D end-to-end models, and the degree of improvement over previous approaches is quantitatively evaluated.

\subsection{Research Gap \& Summary}
Recently, there has been growing interest in using generative models such as GANs and diffusion models to create novel 2D design images in the field of automotive wheel design. These 2D designs are then either converted into 3D models through simple extrusion techniques or used to estimate 3D geometry directly from images in real-time. However, such approaches have clear limitations.

First, simple 2D–3D conversion methods struggle to realistically reproduce complex geometric details, such as spoke structures and curved rims, essential to wheel design.
Second, 3D models generated directly from 2D images often lack design consistency and suffer from poor mesh quality, making them unsuitable for editing or further analysis. As a result, these models fail to meet the precision required for structural simulations, making it difficult to evaluate engineering performance metrics such as mass, stiffness, and natural frequency reliably.
Third, although domain-specific 3D datasets for engineering design and simulation have recently become more widespread, there remains a lack of publicly available, high-quality datasets for components like automotive wheels, which exhibit complex geometries and fine structural details.  
Such a lack of domain-specific datasets makes it challenging to apply AI-based design processes to wheel design effectively.
This study addresses this gap by proposing a practical data generation framework tailored to the wheel design domain.

\section{Methodology}

\subsection{Overall Framework}

\begin{figure}[!ht]
    \centering
    \includegraphics[width=1\linewidth]{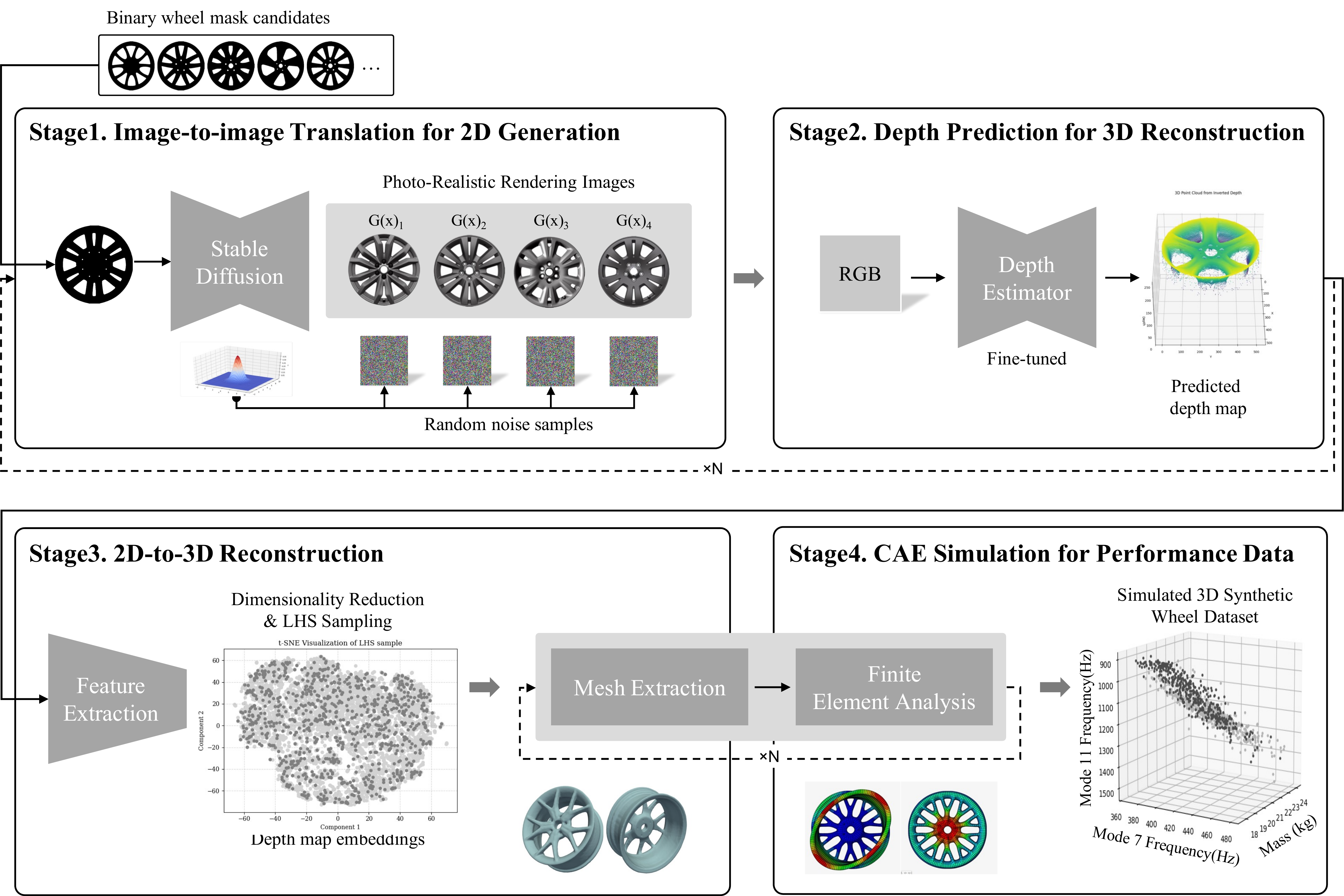}
    \caption{Research framework}
    \label{Research framework}
\end{figure}

This study presents an integrated design framework that converts 2D designs into 3D shapes and evaluates their structural performance. The framework consists of four sequentially connected stages, as illustrated in 
\autoref{Research framework}.

\begin{enumerate}
\item \textbf{Stage 1: Image-to-image Translation for 2D Generation}
\begin{itemize}
\item Generate rendering images using a Stable Diffusion-based image-to-image translation technique
\item Create 6,249 variations by applying different structures and random seed values
\end{itemize}

\item \textbf{Stage 2: Depth Prediction for 3D Reconstruction}
\begin{itemize}
\item Fine-tune a depth prediction model using domain-specific RGB-D data
\item Predict depth maps from 2D renderings to generate 2.5D representations (6,249 in total)
\end{itemize}

\item \textbf{Stage 3: 2D-to-3D Reconstruction}
\begin{itemize}
\item Perform dimensionality reduction on depth maps to sample representative designs
\item Reconstruct selected samples into complete 3D mesh models (n=904)
\end{itemize}

\item \textbf{Stage 4: CAE Simulation for Performance Evaluation}
    \begin{itemize}
    \item Conduct structural analysis simulations on reconstructed 3D mesh models
    \item Generate a structural analysis dataset from simulation outputs
    \end{itemize}
\end{enumerate}

The proposed framework enables the automated generation and structural evaluation of high-quality 3D designs, supporting efficient and scalable exploration of the design space.

\subsection{Stage1: Image to Image Translation for 2D Generation}

\begin{figure}[!htbp]
    \centering
    \includegraphics[width=1\linewidth]{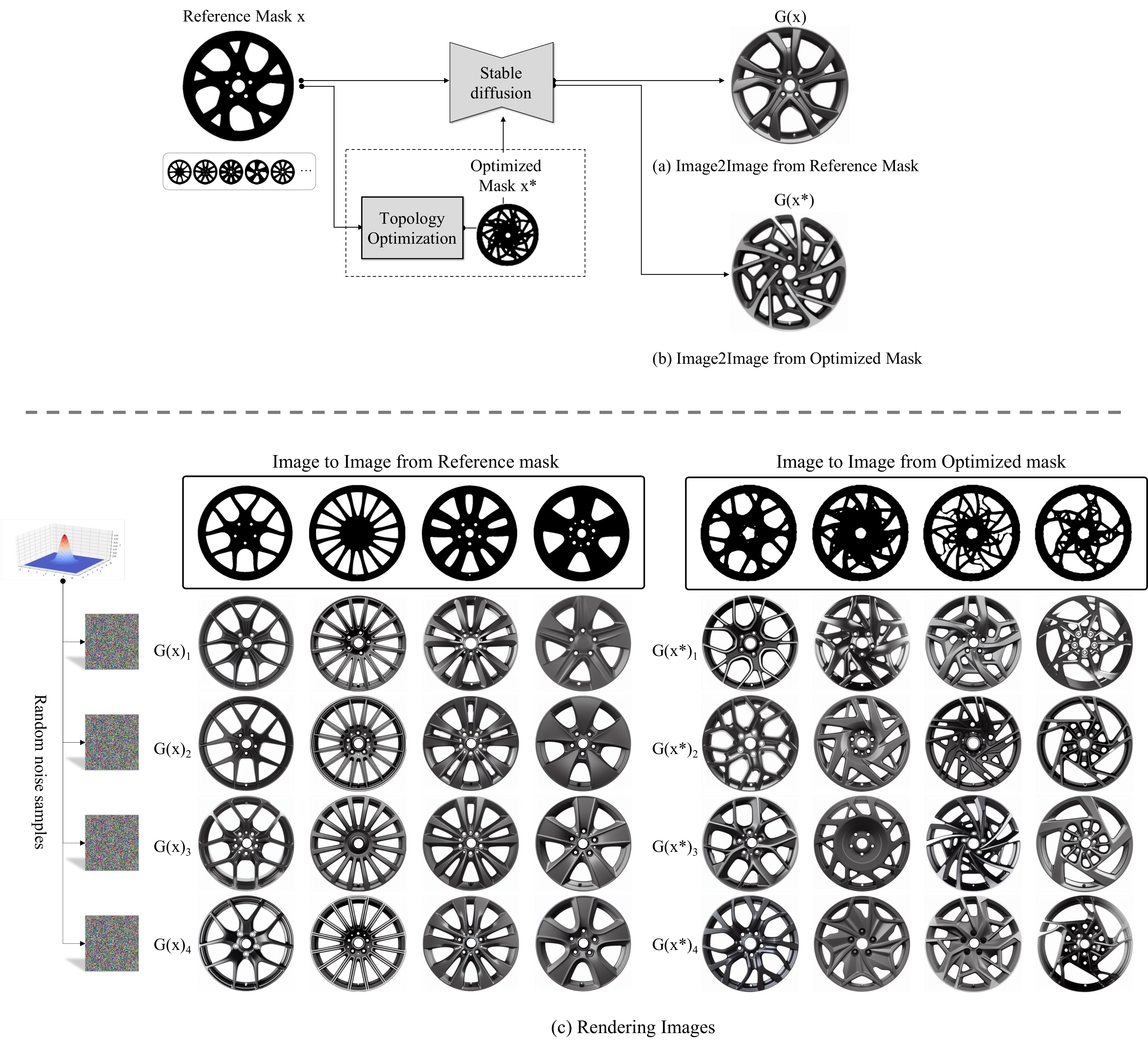}
    \caption{Overall workflow of generating diverse 2D wheel designs and photorealistic renderings.}
    \label{2d_synthetic_designs}
\end{figure}

Combining these techniques generates creative structural designs from reference images using topology optimization, and diverse renderings are produced using Stable Diffusion.

\subsubsection{Design Variations using Topology Optimization}

This stage adopts topology optimization for wheel design, following the method proposed by Oh et al. (2019) \cite{oh2019}. In their formulation, topology optimization maximizes structural performance by optimizing the material distribution under prescribed loads and boundary conditions. The problem is defined as a multi-objective optimization that minimizes structural compliance while maintaining similarity to a reference design:

\begin{equation}
    \begin{aligned}
    \text{min } f(\mathbf{x}) &= \mathbf{U}^{T}\mathbf{K(x)}\mathbf{U} + \lambda\|\mathbf{x}_r-\mathbf{x}\|_{1} \\
    \text{s.t.}\quad & \frac{V(\mathbf{x})}{V_0}=f \\
    & \mathbf{K}\mathbf{U} = \mathbf{F} \\
    & 0 \leq x_e \leq 1,\quad e=1,\dots,N_e
    \end{aligned}
    \label{Topology}
\end{equation}

The design variable vector \(\mathbf{x} \in \mathbb{R}^{N_e}\) represents the material density of each element, where 0 and 1 indicate void and solid, respectively. \(\mathbf{U}\) is the displacement vector, \(\mathbf{K(x)}\) is the global stiffness matrix dependent on the material distribution, and \(\mathbf{F}\) is the external load vector. The first term in the objective function measures compliance, which is minimized to enhance structural stiffness. The second term enforces similarity to the reference design \(\mathbf{x}_r\), weighted by the factor \(\lambda\). The constraints ensure that the material usage satisfies the prescribed volume fraction \(f\), and that element densities remain within the range \([0, 1]\). In this formulation, \(V(\mathbf{x})\) denotes the volume of the material used in the design, \(V_0\) is the volume of the entire design domain, and \(N_e\) is the total number of finite elements.

In addition to the original formulation, the design domain is constrained to a single segment to ensure rotational symmetry. The full wheel design is then completed by duplicating the optimized segment four, five, or six times. A variety of topology-optimized wheel designs are generated by varying the four parameters: the similarity weight \(\lambda\), the ratio of normal to shear loads, the volume fraction \(f\), and the symmetry parameter \(n_{\text{seg}}\).

For detailed information on generating topology-optimized wheels, please refer to the work by Oh et al. (2019)\cite{oh2019}.

\subsubsection{Photorealistic Rendering based on Stable Diffusion}

A pre-trained Stable Diffusion model~\cite{Rombach2022} renders realistic images from base designs through text-conditioned image-to-image translation. The inference process begins by encoding the input image \( \mathbf{x}_0 \) into a latent representation \( \mathbf{z}_0 = \mathcal{E}(\mathbf{x}_0) \) using a Variational Autoencoder (VAE) encoder \( \mathcal{E} \). Subsequently, Gaussian noise is added to \( \mathbf{z}_0 \) according to a 'strength' parameter, resulting in a noisy latent representation \( \mathbf{z}_t \). This 'strength' parameter modulates the degree to which the original image structure is preserved. The noisy latent representation \( \mathbf{z}_t \) is then iteratively refined using a U-Net based noise predictor \( \epsilon_\theta \), conditioned on the text embedding \( \tau_y \). This model \( \epsilon_\theta \) predicts and removes the added noise at each step, progressively enhancing the latent representation through the reverse process. The Classifier-Free Guidance (CFG)~\cite{ho2022classifierfree} method is employed to control the influence of the text condition by combining conditional and unconditional predictions as follows:
\begin{equation}
\tilde{\epsilon}_\theta(\mathbf{z}_t, \tau_y) = (1 + \gamma)\epsilon_\theta(\mathbf{z}_t, t, \tau_y) - \gamma \epsilon_\theta(\mathbf{z}_t, t, \emptyset)
\end{equation}
where \( \gamma \) denotes the guidance scale, controlling the strength of the text conditioning, and \( \emptyset \) represents the unconditional (null prompt) case. Upon completion of the denoising process, the refined latent representation \( \mathbf{z}_0' \) is decoded into the final output image \( \mathbf{x}_{\text{output}} = \mathcal{D}(\mathbf{z}_0') \) by the VAE decoder \( \mathcal{D} \).

For the experiments, the LCMScheduler~\cite{Salmon2024} was employed for efficient sampling. The 'strength' parameter was set to 0.75 (typical values range from 0.5 to 0.8, with higher values resulting in more significant deviation from the original image). The guidance scale \( \gamma \) was set to 7.5 (typically ranging from 5 to 15, controlling the adherence to the text prompt), and the number of sampling steps was specified as 50.

\subsubsection{Overall 2D Design Workflow}
The overall workflow for generating photorealistic 2D designs involved using both reference masks and masks derived from topology optimization as starting points for image generation. These masks served as the basis for the input image condition \( \mathbf{x}_0 \) in the Stable Diffusion image-to-image process. For each mask, the text prompt \textit{'High-performance car wheel rim, detailed 3D rendering'} was utilized to guide the creation of realistic images. Applying this method, along with incorporating variations through different seed values, resulted in the generation of 6,249 final rendered images. Each final image possesses a resolution of \(512 \times 512 \times 3\) in RGB format.

\autoref{2d_synthetic_designs} visually summarizes key aspects of this workflow. Specifically, ~\ref{2d_synthetic_designs}(a) shows an example rendered directly from a reference mask, ~\ref{2d_synthetic_designs}(b) displays a result derived from a topology-optimized mask, and ~\ref{2d_synthetic_designs}(c) illustrates how different designs can be created from the same mask by changing the seed value. This integrated workflow allowed for the efficient exploration of many diverse design ideas, significantly expanding the range of design possibilities. 

\subsection{Stage2: Depth Prediction for 3D Reconstruction}
Predicting depth from single monocular images is an inherently ill-posed problem. Additionally, obtaining large-scale ground truth depth data is difficult, especially for specific objects such as automotive wheels. To address these challenges, this study employed the Marigold model~\cite{Marigold2024}, a foundation model known for its strong generalized depth prediction performance. The pre-trained Marigold model was fine-tuned using a small, synthetic RGB-D dataset generated from CAD models, targeting the specific problem of automotive wheel depth prediction.

\subsubsection{Depth Prediction Foundation Model}
This section details the specific generative formulation, network architecture, adaptations to the U-Net, and inference process as presented in the original Marigold study~\cite{Marigold2024}.

\begin{figure}[!ht]
\centering
\includegraphics[width=1\linewidth]{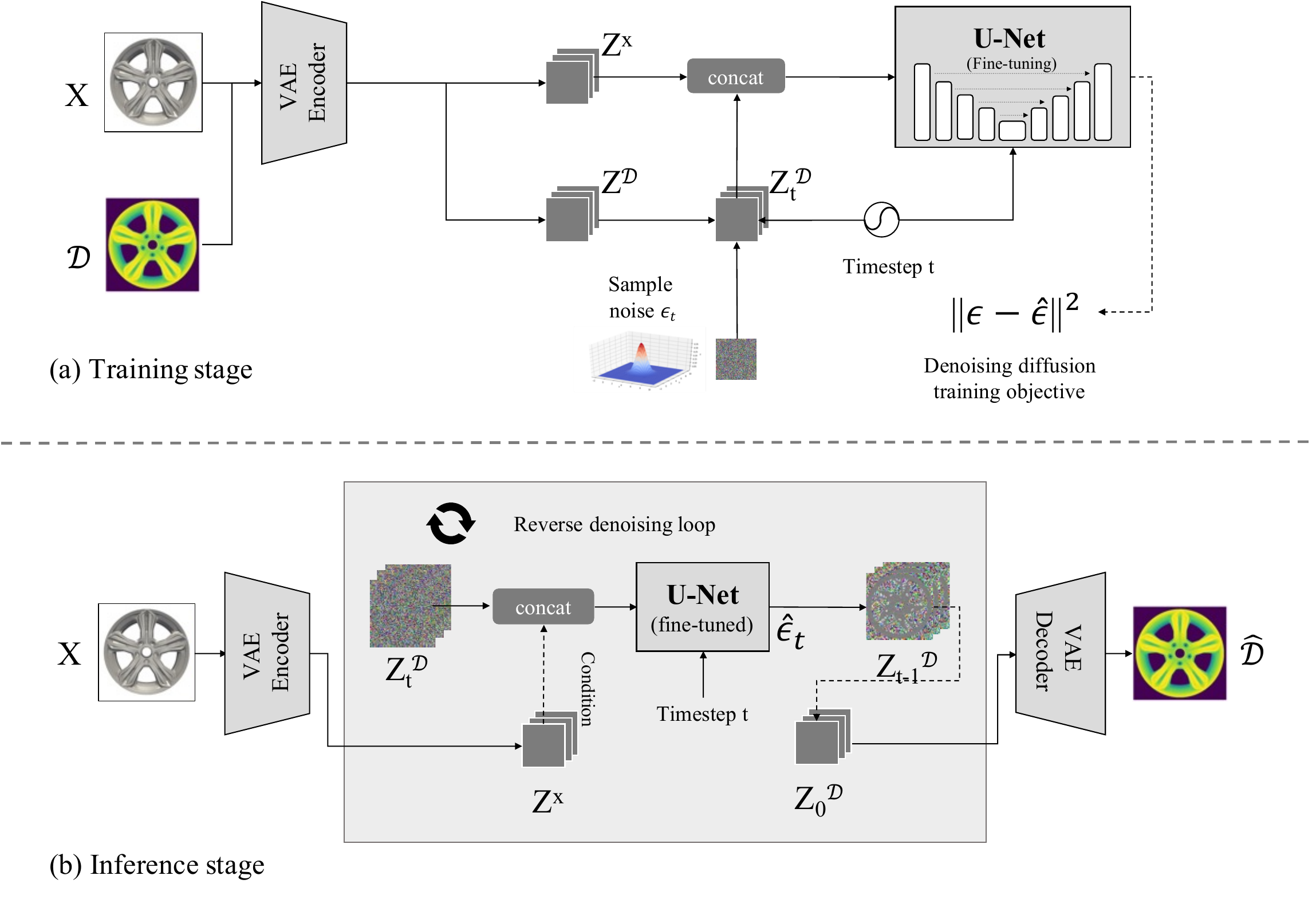}
\caption{Architecture of depth estimator(Marigold)}
\label{Marigold_pipeline}
\end{figure}

\paragraph{Generative Formulation}
The Marigold model formulates monocular depth prediction as a conditional diffusion generation problem. Given an RGB image \( \mathbf{x} \in \mathbb{R}^{W \times H \times 3} \) as a condition, it models the conditional distribution \( D(\mathbf{d} | \mathbf{x}) \) of the depth map \( \mathbf{d} \in \mathbb{R}^{W \times H} \). In the forward process, Gaussian noise is progressively added to the clean depth map, as follows:

\begin{equation}
    \mathbf{d}_t = \sqrt{\bar{\alpha}_t}\mathbf{d}_0 + \sqrt{1 - \bar{\alpha}_t}\boldsymbol{\epsilon}, \quad \boldsymbol{\epsilon} \sim \mathcal{N}(0, I)
    \label{eq:forward_diffusion}
\end{equation}

Here, \( \bar{\alpha}_t = \prod_{s=1}^t (1 - \beta_s) \), where \( t \in \{1, ..., T\} \), and \( \beta_s \) is the noise variance schedule. In the reverse process, the denoising model \( \boldsymbol{\epsilon}_\theta(\mathbf{d}_t, \mathbf{x}, t) \), parameterized by \( \theta \), is trained to gradually remove the noise and recover \( \mathbf{d}_{t-1} \) from \( \mathbf{d}_t \).

During training, a random timestep \( t \) is selected to generate \( \mathbf{d}_t \) by adding noise to \( \mathbf{d}_0 \), and the model learns to predict the noise \( \hat{\boldsymbol{\epsilon}} = \boldsymbol{\epsilon}_\theta(\mathbf{d}_t, \mathbf{x}, t) \). The objective is to minimize the difference between the predicted and true noise via the following loss:

\begin{equation}
    \mathcal{L} = \mathbb{E}_{\mathbf{d}_0, t, \boldsymbol{\epsilon}}\left[ \| \boldsymbol{\epsilon} - \boldsymbol{\epsilon}_\theta(\mathbf{d}_t, \mathbf{x}, t) \|^2 \right]
    \label{eq:ddpm_loss}
\end{equation}

\paragraph{Network Architecture}
The network of Marigold is illustrated in \autoref{Marigold_pipeline}. The model is based on Stable Diffusion v2, a pre-trained latent diffusion model (LDM) originally trained on large-scale text-image pairs from the LAION-5B dataset. With minimal modifications, this model is adapted for image-conditional depth prediction.

The input RGB image \( \mathbf{x} \) and its corresponding depth map \( \mathbf{d} \) are encoded into the latent space using a fixed VAE encoder \( \mathcal{E} \), resulting in latent vectors \( \mathbf{z}^{(x)} = \mathcal{E}(\mathbf{x}) \) and \( \mathbf{z}^{(d)} = \mathcal{E}(\mathbf{d}) \). Since the depth map is single-channel, it is duplicated to three channels before being passed into the VAE. The decoded output is then averaged across channels to reconstruct the final depth map.

\paragraph{Adapted Denoising U-Net}
The conditional denoising model \( \boldsymbol{\epsilon}_\theta \) takes the latent image \( \mathbf{z}^{(x)} \) and the noise-added latent depth \( \mathbf{z}_t^{(d)} \), and concatenates them along the channel dimension as input:

\begin{equation}
    \mathbf{z}_t = \text{Concat}(\mathbf{z}^{(d)}_t, \mathbf{z}^{(x)})
    \label{eq:latent_concat}
\end{equation}

This modification effectively doubles the number of input channels to the U-Net. To maintain stable activation magnitudes in the first layer, the original weights are duplicated and scaled by a factor of 0.5 during initialization, ensuring a smooth transition in network behavior.

\paragraph{Inference}
During inference, the input image is encoded into \( \mathbf{z}^{(x)} \) using the VAE encoder, and the latent depth vector \( \mathbf{z}^{(d)}_T \) is initialized with standard Gaussian noise. Then, using the DDIM-based non-Markovian sampling procedure, the latent vector is progressively denoised from \( \mathbf{z}^{(d)}_T \) to \( \mathbf{z}^{(d)}_0 \). Finally, the latent vector is decoded via the VAE decoder to generate the final depth map. For optimal balance between speed and quality, 10 denoising steps are used during inference in this study. Detailed implementation can be found in~\cite{Marigold2024}.

\subsubsection{RGB-D data augmentation method}

\begin{figure}[h!]
    \centering
    \includegraphics[width=.7\linewidth]{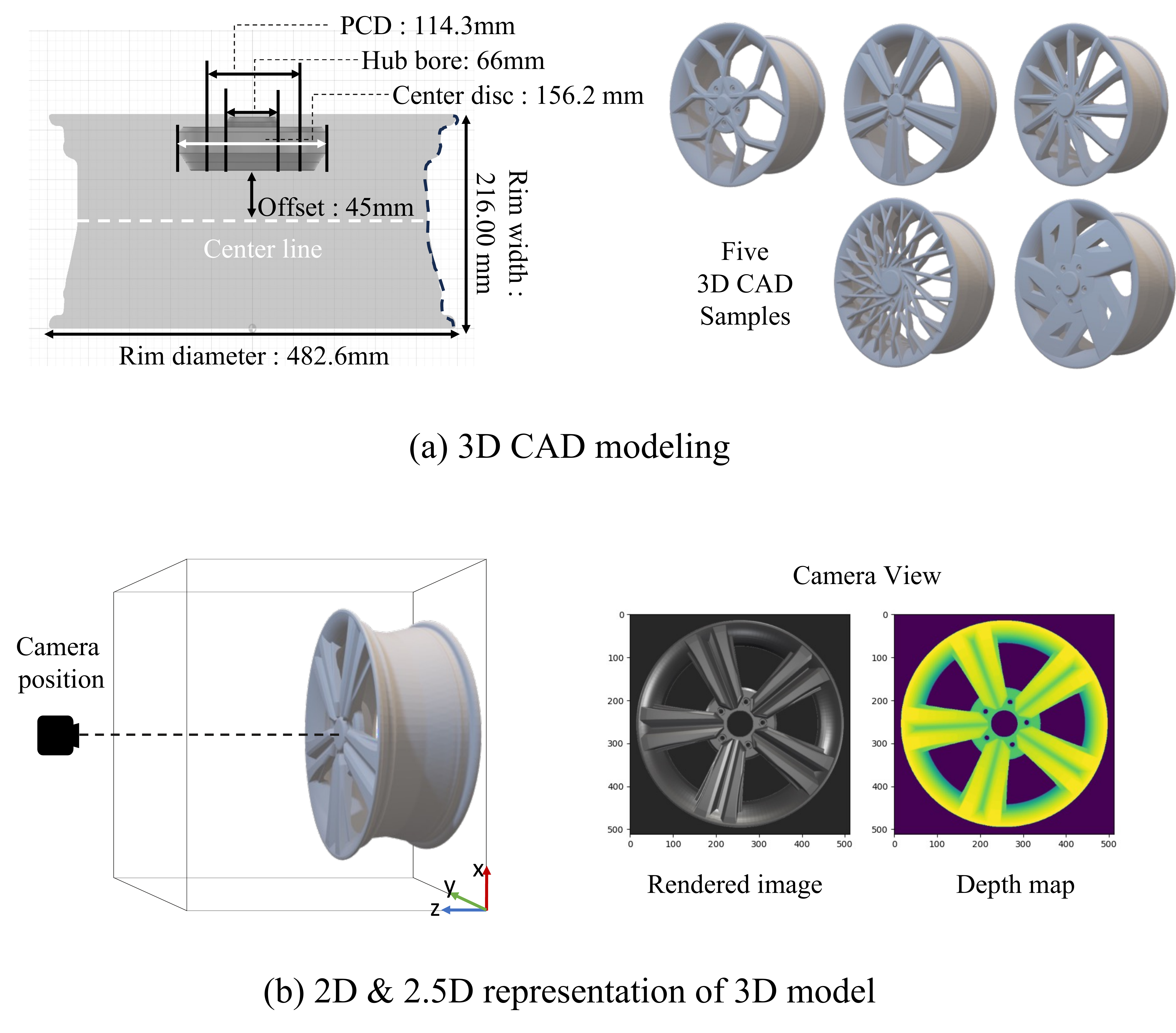}
    \caption{RGB-D dataset collection}
    \label{RGB-D Colletion}
\end{figure}

\begin{figure}[h!]
    \centering
    \includegraphics[width=1\linewidth]{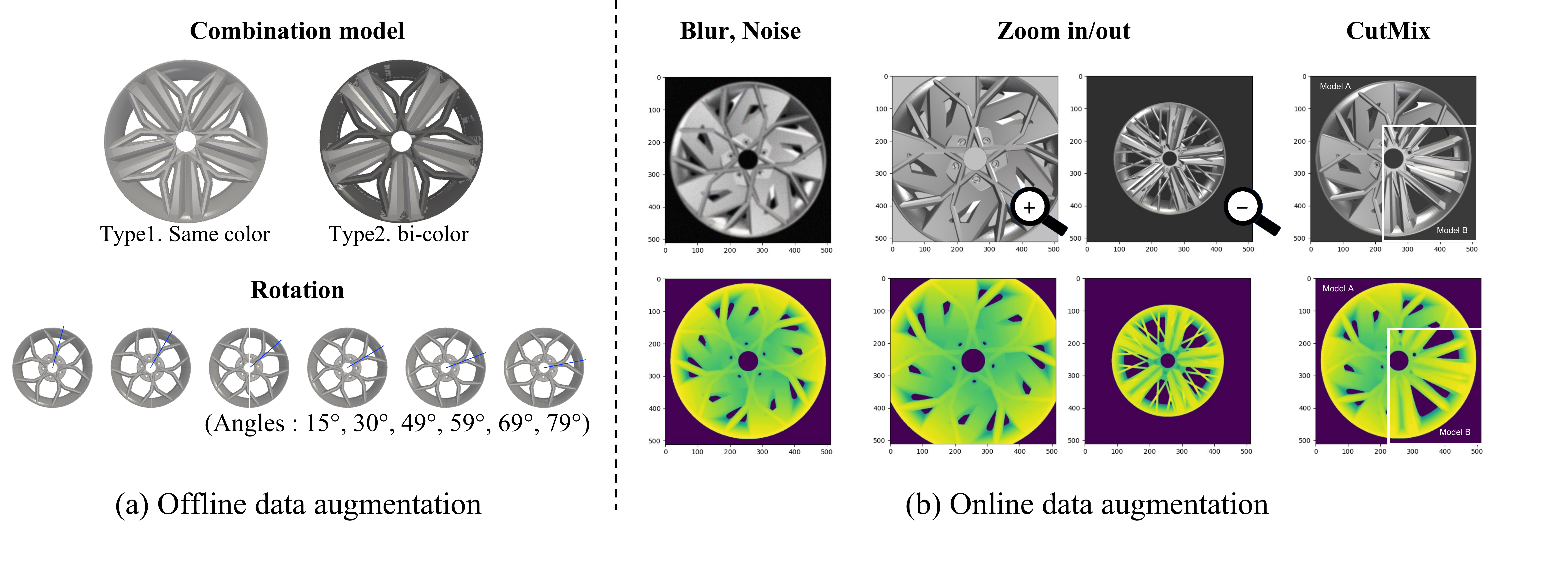}
    \caption{RGB-D data augmentation}
    \label{RGB-D Data Augmentation}
\end{figure}

This study constructed an RGB-D dataset using a limited number of 3D CAD wheel models. 3D CAD modeling was conducted for five distinct wheel designs. The CAD designs were created based on fundamental wheel components, referencing industry-standard specifications from relevant websites~\cite{wheel_size}. According to these specifications, the following parameters were used:  a rim diameter of 482.6 mm(19 inches), a rim width of 216 mm(8.5 inches), an offset of 45 mm, a pitch circle diameter (PCD) of 114.3 mm, a hub bore diameter of 66 mm and a center disc diameter of 156.2 mm (~\ref{RGB-D Colletion}(a)). For each handcrafted 3D CAD model, an RGB image and corresponding depth maps were captured from a frontal perspective, resulting in both 2D and 2.5D representations of the models (~\ref{RGB-D Colletion}(b)).

Offline and online data augmentation strategies were implemented to learn richer data representations(\autoref{RGB-D Data Augmentation}. In the offline augmentation phase(~\ref{RGB-D Data Augmentation}(a)), datasets were created by combining two-wheel models. These were structured in two ways: Type 1, where RGB image variations were set with identical colors for each model, and Type 2, where different colors were applied. Additionally, diverse frontal views were captured by rotating each model around the z-axis at angles of 15°, 30°, 49°, 59°, 69°, and 79°.
Online augmentation(~\ref{RGB-D Data Augmentation}(b)) was performed randomly during training in real-time, applying various effects such as noise, blur, zoom effects, and CutMix (which combines half of each image from two models). These data augmentation techniques made it possible to build a rich dataset from a few wheel models and train the system to predict various features and environmental conditions.

\subsection{Stage3: 2D to 3D Reconstruction}
\autoref{2D-to-3D pipeline} shows an overview of the entire process of reconstructing the rim and spokes of a wheel into a three-dimensional model using 2D images. This section explains the pipeline process.

\subsubsection{2D-to-3D Pipeline}
First, the 2D input image is converted into depth information through the depth prediction model described earlier and transforms the spoke geometry into a point cloud. In the spoke processing step, unnecessary information is removed from the spoke area to use only valid spoke regions. In the rim processing step, the rim shape is reconstructed using a predefined reference mesh. During the rim processing, the lower surface, including the center disc and the rim's outer surface is modeled as point clouds. (See 2D-to-3D pipeline (a)).

\begin{figure}[h!] 
\centering 
\includegraphics[width=1\linewidth]{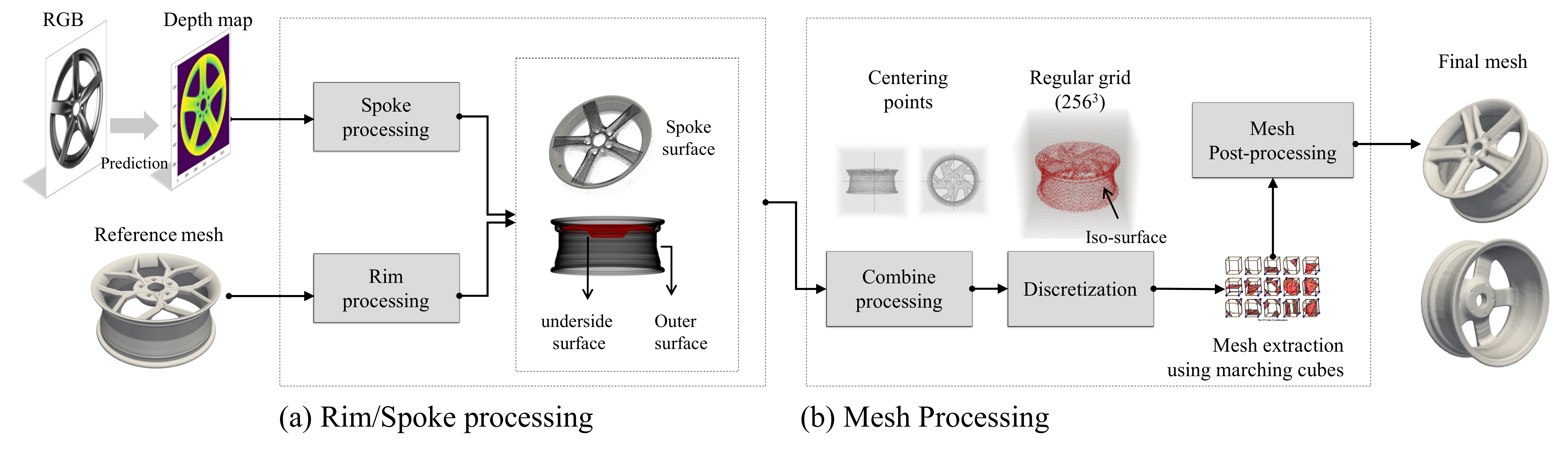} 
\caption{2D-to-3D reconstruction pipeline} 
\label{2D-to-3D pipeline} 
\end{figure}

After individually processing the rim and spokes, they are integrated into a complete wheel shape. First, each element extracted as a point cloud is aligned for the spatial center point, and then the point data is discretized in a regular grid space. Next, the marching cubes algorithm is applied to extract the isosurface at the boundary of the value, forming a mesh for the 3D surface. Finally, the mesh post-processing process completes the watertight 3D wheel mesh by smoothing the mesh, reducing the number of polygons, and removing unnecessary artifacts (~\ref{2D-to-3D pipeline}(b)).

\begin{figure}[htbp!] 
\centering 
\includegraphics[width=1\linewidth]{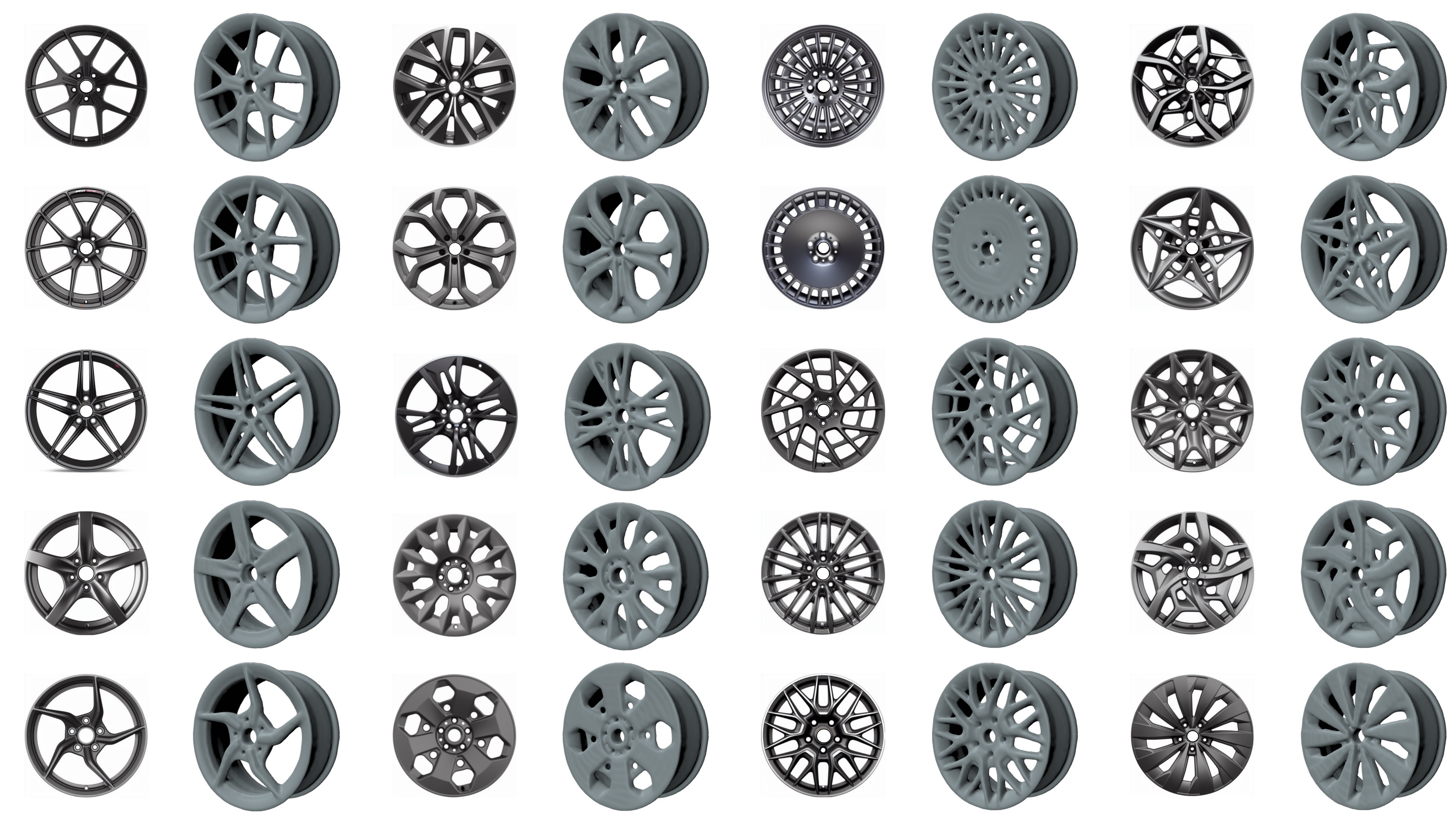} 
\caption{Results of 2D-to-3D wheel reconstruction} 
\label{fig:2d_to_3d_examples} 
\end{figure}

\begin{figure}[htbp!]
    \centering
    \includegraphics[width=1\linewidth]{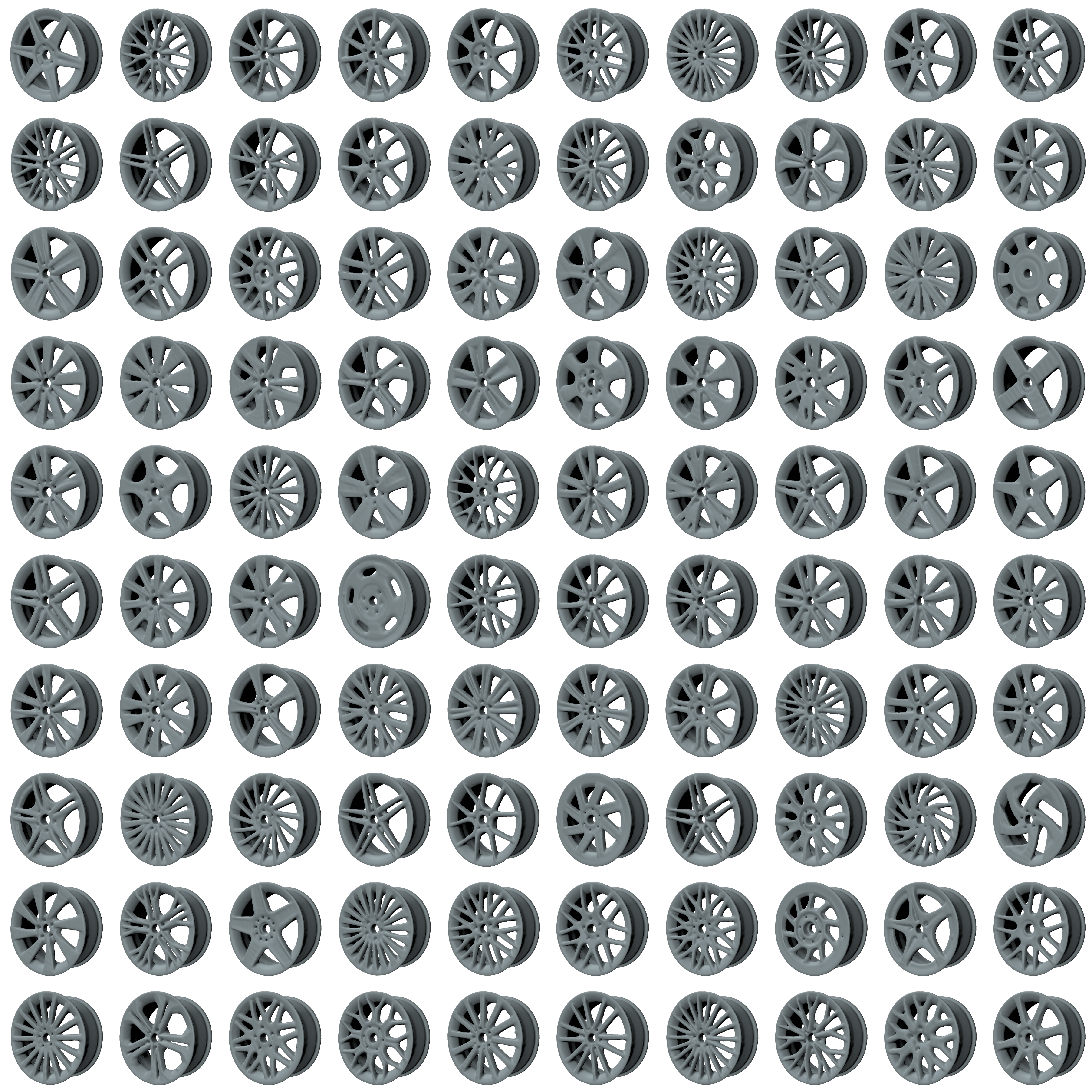}
    \caption{Large-scale visualization of reconstructed 3D wheel} 
    \label{fig:large_scale_dataset}
\end{figure}

The proposed method effectively combines depth maps from 2D images with reference meshes to reconstruct realistic 3D wheel models. \autoref{fig:2d_to_3d_examples} demonstrates the reconstruction results from the proposed pipeline. The figure presents 2D images with various spoke designs on the left and their corresponding 3D wheel models on the right. It shows accurate reproduction of complex features, including intricate spoke patterns and internal rim surfaces.

\autoref{fig:large_scale_dataset} illustrates a grid of 100 3D wheel meshes generated by the proposed algorithm. These results, featuring diverse spoke configurations, confirm that the framework achieves design variety and structural consistency. The reconstruction process requires approximately 10 seconds for point cloud generation and 1 minute for high-resolution mesh conversion. Processing times are expected to decrease with future computational optimizations. The resulting meshes immediately apply to product visualization, simulation, 3D printing, and related applications.

\subsection{Stage4: CAE Simulation for Performance Data}

\subsubsection{Design Sampling via LHS}

This study utilized approximately 6,000 pre-rendered wheel images generated by a Stable Diffusion-based image generation model to obtain diverse wheel shapes. These rendered images contain high-dimensional visual information such as light sources, reflections, and surface textures. Applying simple dimensionality reduction directly to rendered images could capture irrelevant visual effects rather than actual wheel geometry. Instead, depth maps were used to focus on structural information. This approach enables more efficient design sampling compared to RGB image-based dimensionality reduction. The implementation process involved three sequential steps: First, depth maps were extracted from each wheel image. Second, these depth maps were transformed into compact feature vectors using Vision Transformer (ViT). Finally, t-SNE created a two-dimensional representation of the feature space. From this reduced space, 1,000 diverse samples were selected using Latin Hypercube Sampling (LHS). This systematic approach eliminates redundant examples while allowing for a wide range of designs to be included.

\subsubsection{Simulation Data Generation}

\paragraph{Simulation Setup}

The wheel design sample was first converted to an initial mesh through a 2D-to-3D reconstruction pipeline. It was then meshed using Rhino Grasshopper~\cite{rhino3d} based on quad elements and converted to Brep (STEP) format. Although some resolution loss may occur during the conversion to the BREP format, this ensures high compatibility with subsequent analysis software and simplifies the pre-processing process. 

The converted model was analyzed in the Altair SimLab~\cite{simlab} environment using normal mode analysis with free-free boundary conditions. 
In particular, Python scripts were used to automate the analysis, processing the entire process from loading the STEP file to creating the volume mesh, applying the properties, setting the analysis options, running the OptiStruct solver~\cite{optistruct}, and saving the results in a batch. This enabled the analysis of many design samples quickly and consistently. 

In the volume mesh generation stage, a 10-node Tetra Element (Tet10) was employed, and the Average Element Size was set to 6 mm to ensure the accuracy of the analysis while maintaining the basic element quality. In addition, the material properties of all analysis models were set to A356-T6 aluminum alloy to reflect the actual manufacturing environment of the wheel. A356-T6 is a cast aluminum alloy widely used in the production of automobile wheels. The applied physical properties are: density 2,680 kg/m³, elastic modulus 72 GPa, Poisson's ratio 0.33, yield strength 175 MPa, and ultimate tensile strength 250 MPa.

\paragraph{Modal Analysis}

\begin{figure}[!ht]
    \centering
    \includegraphics[width=.7\linewidth]{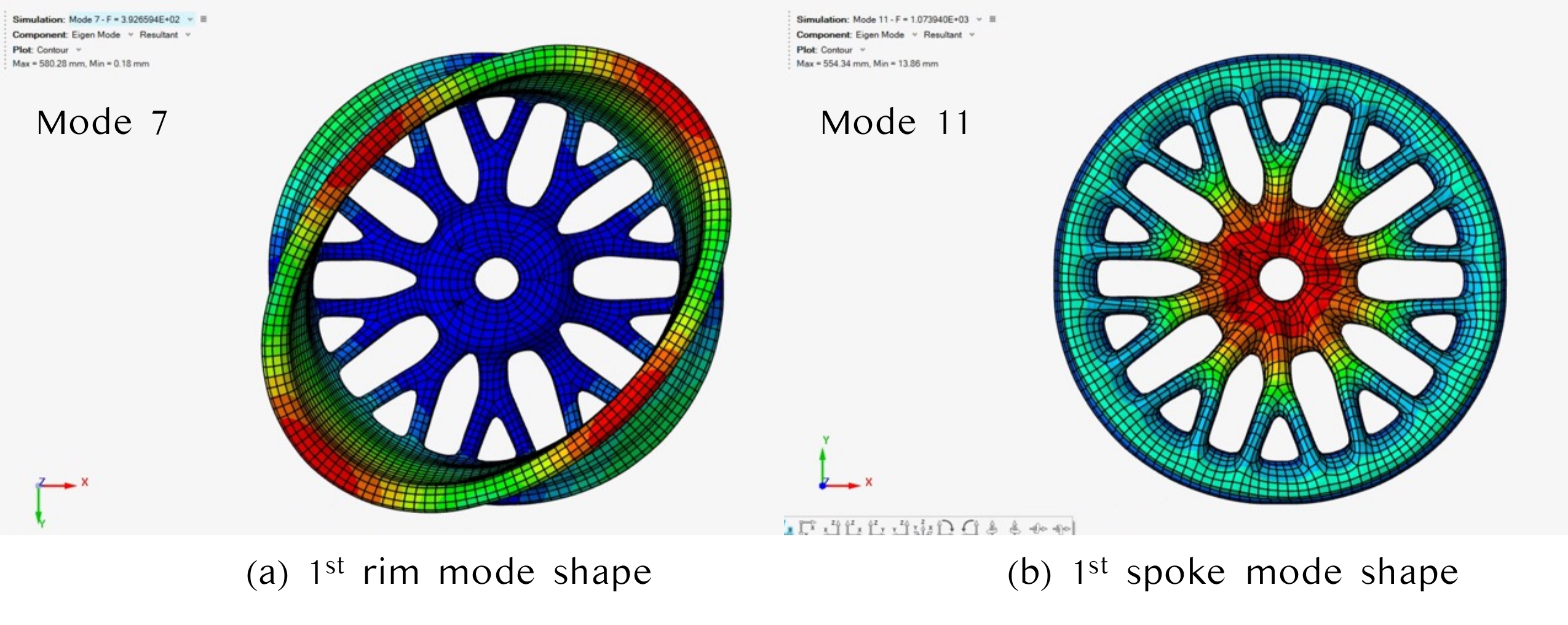}
    \caption{modal analysis of reconstructed 3D model}
    \label{Mode Analysis}
\end{figure}

\autoref{Mode Analysis} presents the results of a modal analysis under free-free boundary conditions, illustrating the natural frequencies and corresponding mode shapes.
Modal analysis results showed that the first and 15th modes were identified, and the first frequencies of the rim mode and spoke mode were seventh (rim mode) and 11th (spoke mode), respectively. In modal analysis, the rim mode (deformation around the rim) and the spoke mode (deformation around the spoke) are the most commonly observed and representative modes that significantly impact driving noise and vibration. These modes concentrate deformation in the rim and spoke areas, respectively, and structural fatigue and NVH performance degradation can be prominent when resonance occurs. In particular, it is efficient to prioritize improving these two modes to maximize the effect of vibration and noise suppression from limited design resources, as even a slight change in the thickness and shape of the rim or spokes can significantly alter their natural frequencies.

Additionally, to verify the consistency of the interpretation results, geometric consistency was analyzed by calculating the centroid based on the depth map of each output for all successfully simulated design samples. A depth map with a resolution of 512×512 was used to calculate the centroid, and only the x and y coordinates of the centroid were considered to evaluate geometric symmetry and balance. The analysis showed that the average center point calculated for all samples was (256.51, 256.22), which is very close to the center of the image (256, 256), indicating that the generated shapes are evenly distributed on the screen. In addition, the dispersion of the center points is low, at 2.64 on the x-axis and 0.71 on the y-axis, indicating that the position deviation between samples is slight. 

These results quantitatively support that the proposed simulation-based generation pipeline maintains a high geometric stability and balance level. In this study, the natural frequencies(7th and 11th modes) and the mass value were added as a key performance score and organized into a dataset.

\section{Results \& Discussion}

\subsection{Depth Estimator Implementation} 

This section describes the fine-tuning implementation of the depth prediction model, focusing on dataset configuration, training strategy, and performance evaluation.

\subsubsection{Dataset Partition: Training, Validation, and Test Sets} 

The dataset was divided into three sets, training, validation, and test, to evaluate the depth prediction model's performance thoroughly. Each partition was designed to verify the model's generalization capabilities. The training set comprised 140 RGB-D images generated from wheel combination models. Data was created using offline augmentation methods, with various online augmentation techniques applied during training to enhance model performance. The validation set consisted of RGB-D images derived from the same combination model as the training set, with modifications to simulate realistic testing environments. These modifications included noise application, blur effects, and new backgrounds. This approach enabled the evaluation of the model's robustness against previously unseen patterns and noisy environments. The test set contained approximately 35 RGB-D samples obtained from single-wheel models, which were completely excluded from the training process. Similar variations to the validation set were applied to this set. This testing environment verified that the model avoided overfitting to training data while maintaining stable performance on new data. The experimental results are summarized in \autoref{tab:model_performance}. Four levels of data augmentation were compared: Level 0 utilized only original unaugmented data; Level 1 applied Gaussian noise and blur; Level 2 incorporated object size transformations, including zoom effects; and Level 3 additionally implemented the CutMix technique.

\subsubsection{Hyperparameter Configuration and Performance Metrics} 

\paragraph{Hyperparameter Configuration}
Model training was conducted using a single A100 NVIDIA GPU.
Input images were resized to a resolution of \(512 \times 512 \times 3\), and the corresponding depth maps were generated at \(512 \times 512 \times 1\). A batch size of 16 was used for all experiments. A conservative learning rate of \(1 \times 10^{-5}\) was adopted in conjunction with the AdamW optimizer to ensure training stability. Overfitting was addressed by applying a weight decay coefficient of 0.01, and the training process was carried out for 100 epochs. A cosine annealing schedule was employed to reduce the learning rate throughout the training progressively. The diffusion process was governed by a Denoising Diffusion Probabilistic Model (DDPM) noise scheduler configured with 1,000 timesteps.

The training was repeated using fixed random seeds {0, 42, 231} to evaluate the model. Data augmentation strategies were empirically optimized. Gaussian noise was added with standard deviations randomly sampled from the range [0.05, 0.1]. Blur was applied using Gaussian kernels with radii varying from 1 to 3. Zoom transformations were introduced with scale factors between 0.7 and 1.2. The CutMix augmentation technique was also employed with a 50\% probability, utilizing randomly selected secondary images during training.

\paragraph{Performance Metrics} Model performance was quantitatively assessed using standard error metrics (RMSE, AbsRel) and an accuracy-based metric ($\delta$).

Root Mean Squared Error (RMSE) is defined as the square root of the mean squared error between the predicted depth $\hat{d}_i$ and the ground truth depth $d_i$:

\begin{equation}
    \text{RMSE} = \sqrt{\frac{1}{M} \sum_{i=1}^{M} \left( \hat{d}_i - d_i \right)^2}
\end{equation}

Absolute Relative Error (AbsRel) is defined as the mean of the absolute error normalized by the ground truth depth:

\begin{equation}
    \text{AbsRel} = \frac{1}{M} \sum_{i=1}^{M} \frac{|\hat{d}_i - d_i|}{d_i}
\end{equation}

Both metrics are used to evaluate the magnitude of prediction errors and the relative accuracy of the model. In addition, the threshold accuracy ($\delta$), which measures the proportion of predictions that fall within a specific ratio of the ground truth, was also analyzed. This study used the threshold of $\delta < 1.25$ to evaluate model accuracy. The metric is defined as:

\begin{equation}
    \delta_{1.25} = \frac{1}{M} \sum_{i=1}^{M} \mathbf{1}\left( \max\left( \frac{\hat{d}_i}{d_i}, \frac{d_i}{\hat{d}_i} \right) < 1.25 \right)
\end{equation}

Here, \( i \) denotes the index of each pixel where a valid depth value exists. For instance, if \(\hat{d}_i\) is within 1.25 times of \(d_i\) (or vice versa), the prediction is considered accurate for that pixel. \( M \) represents the total number of valid pixels, and \( \mathbf{1}(\cdot) \) is an indicator function that returns 1 if the condition is satisfied. For more details on these metrics' mathematical definitions and descriptions, please refer to \cite{eigen2014depth, Marigold2024}.

\subsubsection{Quantitative Results} 

\begin{figure}[!htbp]
    \centering
    \includegraphics[width=0.8\linewidth]{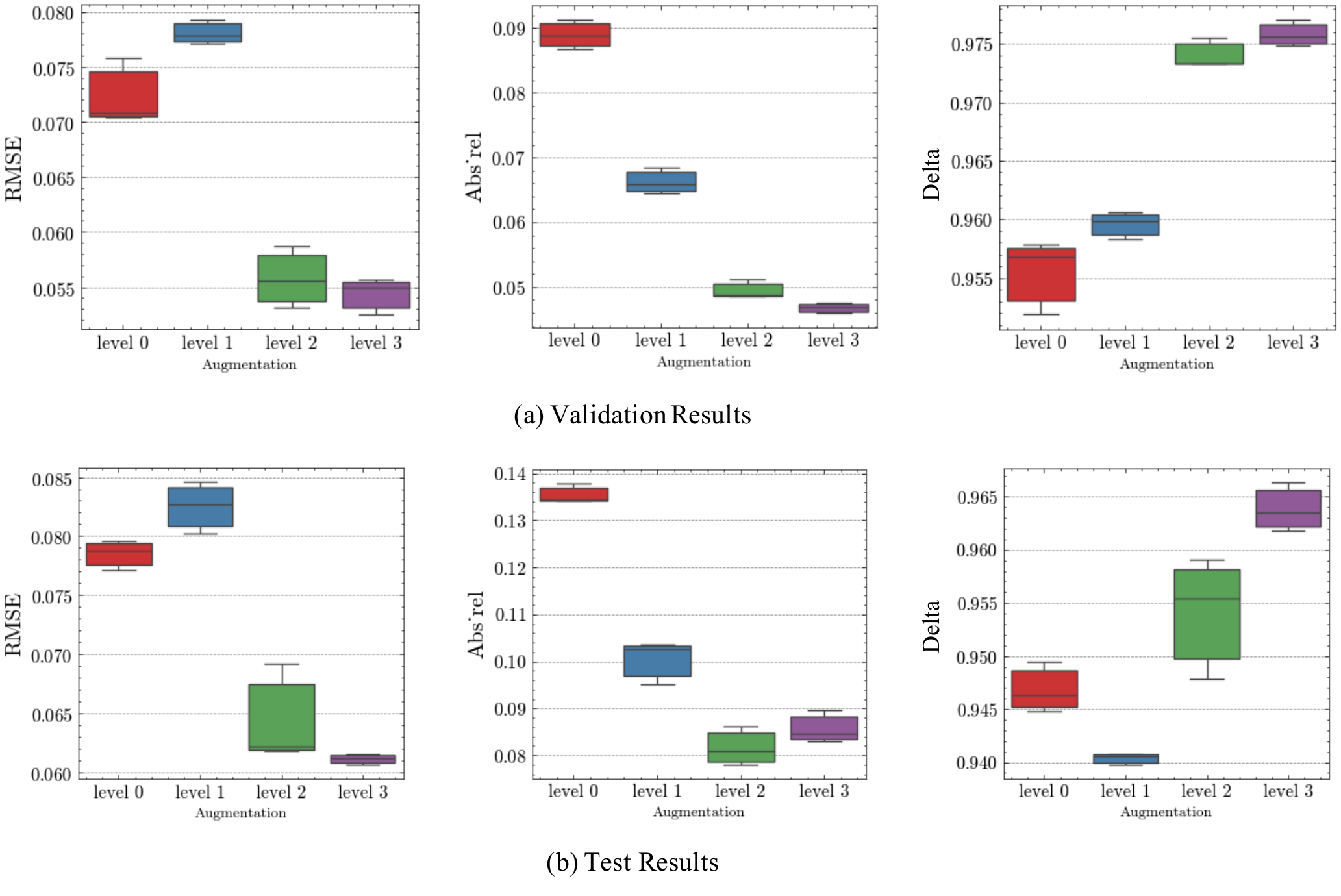}
    \caption{Box plot of model performance metrics (RMSE, AbsRel, \(\delta_{1.25}\)) across different data augmentation levels (Level 0–3)}
    \label{Error Box Plot}
\end{figure}

\autoref{tab:model_performance} and \autoref{Error Box Plot} provide a quantitative analysis of model performance across different data augmentation levels (Level 0–3). The box plots illustrate the distribution and variability of the results across repeated trials, enabling intuitive assessment of each augmentation strategy’s effect.

Validation results show that applying Zoom-In/Out (Level 2) and CutMix (Level 3) significantly reduced RMSE and AbsRel while increasing \(\delta_{1.25}\), indicating both improved prediction accuracy and enhanced stability. In particular, CutMix achieved the lowest RMSE and the highest \(\delta_{1.25}\), demonstrating the best overall performance.

In contrast, Level 1 (noise and blur only) improved AbsRel and \(\delta_{1.25}\) to some extent but led to an increase in RMSE, suggesting that low-level transformations alone may introduce noise detrimental to overall prediction accuracy. Similar trends were observed on the test set, where Zoom-In/Out and CutMix contributed to improved generalization. Notably, CutMix recorded the highest \(\delta_{1.25}\) accuracy; however, in some cases, it showed a slight increase in AbsRel, implying potential errors for specific samples.

Overall, advanced augmentation strategies such as Zoom-In/Out and CutMix significantly enhanced predictive accuracy and generalization. CutMix was particularly effective in improving \(\delta_{1.25}\), though care must be taken in tuning its parameters, as inappropriate region size or placement may disrupt the structural coherence of input images. Meanwhile, the effect of noise and blur was found to vary depending on intensity and kernel size, underscoring the importance of careful augmentation design.

\begin{table}[!htbp]
    \centering
    \caption{Model performance across different augmentation levels and seeds}
    \begin{tabular}{lllccc}
        \toprule
        \textbf{Test Type} & \textbf{Model Type} & \textbf{Seed} & \textbf{RMSE}↓ & \textbf{AbsRel}↓ & \textbf{$\delta_{1.25}$}↑ \\
        \midrule
        \multirow{12}{*}{\makecell{Validation}} 
        & \multirow{3}{*}{Level 0 (no augmentation)} 
            & 0     & 0.076  & 0.091 & 0.952 \\
            &       & 42     & 0.071 & 0.089 & 0.958 \\
            &       & 231    & 0.070 & 0.087 & 0.957 \\
        \cmidrule(lr){2-6}
        & \multirow{3}{*}{Level 1 (noise, blur)} 
            & 0     & 0.079  & 0.065 & 0.958 \\
            &       & 42     & 0.078 & 0.066 & 0.960 \\
            &       & 231    & 0.077 & 0.068 & 0.961 \\
        \cmidrule(lr){2-6}
        & \multirow{3}{*}{Level 2 (noise, blur, zoom)} 
            & 0     & 0.053  & 0.049 & 0.976 \\
            &       & 42     & 0.059 & 0.051 & 0.973 \\
            &       & 231    & 0.056 & 0.049 & 0.973 \\
        \cmidrule(lr){2-6}
        & \multirow{3}{*}{Level 3 (noise, blur, zoom, cutmix)} 
            & 0     & 0.056          & 0.047          & 0.975 \\
            &       & 42     & \textbf{0.053} & \textbf{0.046} & \textbf{0.977} \\
            &       & 231    & 0.055          & 0.047          & 0.976 \\
        \midrule
        \multirow{12}{*}{\makecell{Test}}
        & \multirow{3}{*}{Level 0 (no augmentation)} 
            & 0     & 0.080 & 0.134 & 0.945 \\
            &       & 42    & 0.079 & 0.134 & 0.946 \\
            &       & 231   & 0.077 & 0.138 & 0.949 \\
        \cmidrule(lr){2-6}
        & \multirow{3}{*}{Level 1 (noise, blur)} 
            & 0     & 0.080 & 0.095 & 0.941 \\
            &       & 42    & 0.083 & 0.103 & 0.941 \\
            &       & 231   & 0.085 & 0.104 & 0.940 \\
        \cmidrule(lr){2-6}
        & \multirow{3}{*}{Level 2 (noise, blur, zoom)} 
            & 0     & 0.062 & 0.081          & 0.955 \\
            &       & 42    & 0.069 & 0.086          & 0.948 \\
            &       & 231   & 0.062 & \textbf{0.078} & 0.959 \\
        \cmidrule(lr){2-6}
        & \multirow{3}{*}{Level 3 (noise, blur, zoom, cutmix)} 
            & 0     & 0.062          & 0.083 & 0.964 \\
            &       & 42    & \textbf{0.061} & 0.090 & \textbf{0.966} \\
            &       & 231   & \textbf{0.061} & 0.085 & 0.962 \\
        \bottomrule
    \end{tabular}
    \label{tab:model_performance}
\end{table}

\subsubsection{Qualitative Results}

\begin{figure}[!htbp]
    \centering
    \includegraphics[width=0.8\linewidth]{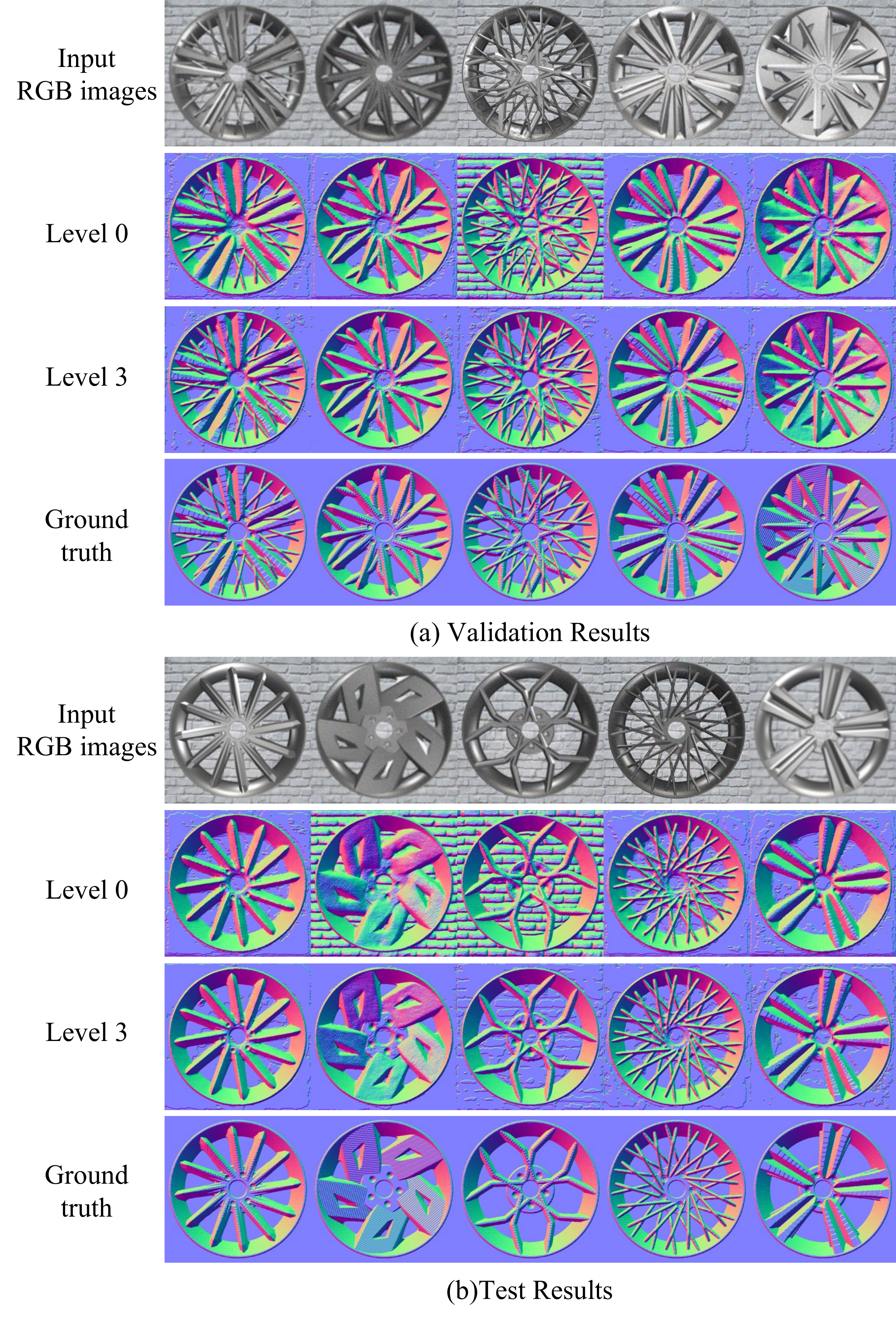}
    \caption{Qualitative results on evaluation \& test dataset}
    \label{Qualitative Results}
\end{figure}

The \autoref{Qualitative Results} visualizes the prediction results for each model as a normal map, which is intended to provide a more precise analysis of the surface geometry and edge representation than a depth map. This allows an intuitive comparison of the prediction's granularity and edge expression. 

~\ref{Qualitative Results}(a) compares the results of the no augmentation (Level 0) and strong augmentation (Level 3: Noise, Blur, Zoom, CutMix applied) models for validation. The Level 0 model had errors in predicting the blurring of complex structures (e.g., spokes) or the inclusion of the background. The Level 3 model showed results close to the correct answer when representing detailed structures.

~\ref{Qualitative Results}(b) shows a similar trend in the Test, and the Level 3 model reliably restores structural features (e.g., spoke boundaries, lug holes) even in new backgrounds and object conditions. This shows that models that have learned various variations can effectively utilize geometric information even in new environments. 

Overall, high-level augmentation techniques such as Level 2 and Level 3 have been shown to improve the accuracy of predictions for complex structures and are particularly effective in restoring boundaries and details precisely, even in unfamiliar environments. Fine-tuning a small number of datasets on the foundation model for depth prediction helps develop domain-specific models. Incorporating data augmentation can further maximize predictive performance.

\subsection{2D-to-3D Reconstruction Evaluation}
\begin{figure}[!ht]
    \centering
    \includegraphics[width=1\linewidth]{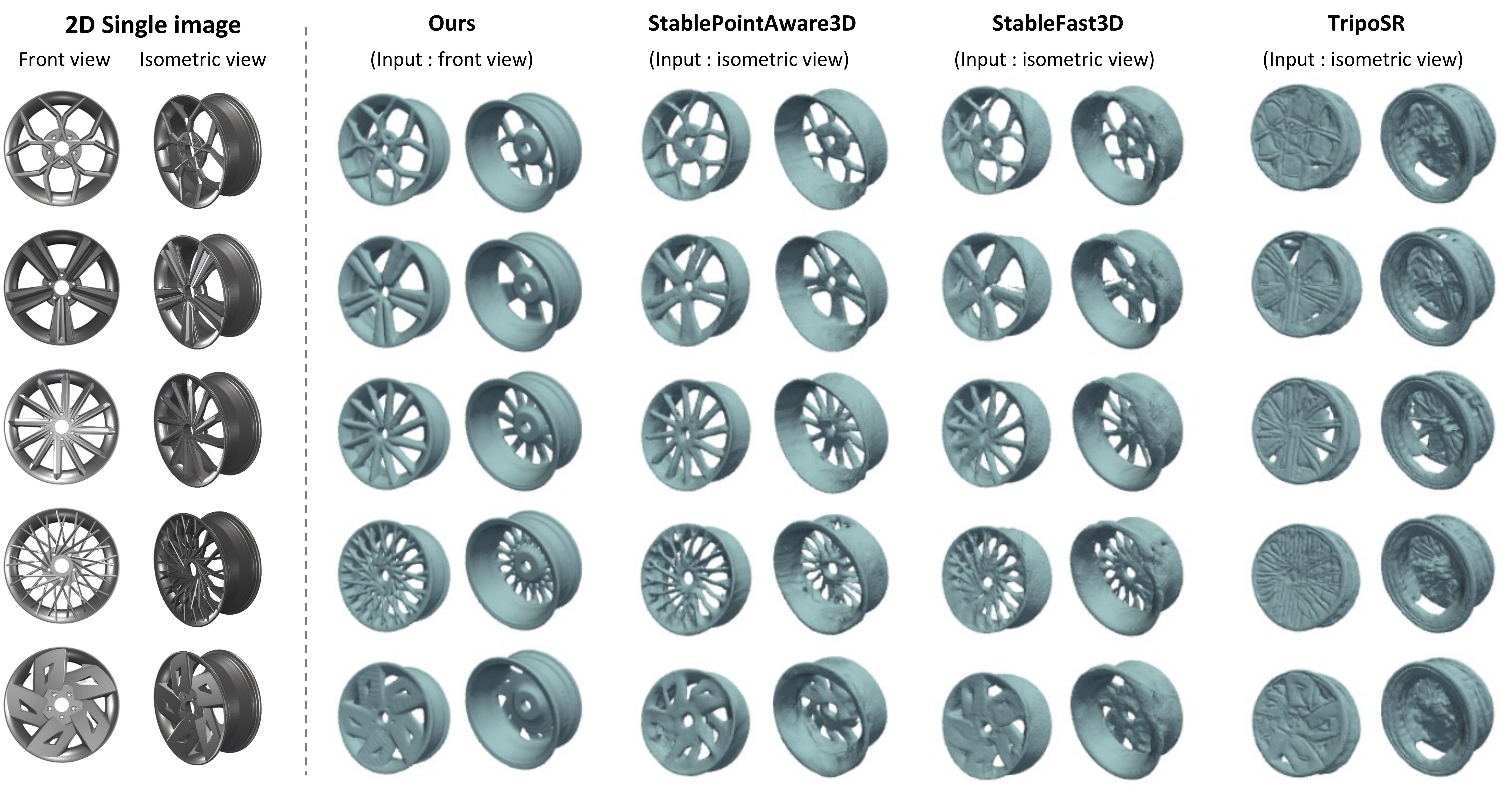}
    \caption{Comparison existing 2D-to-3D approachs}
    \label{Comparison existing 2D-to-3D approachs}
\end{figure}

\begin{table}[!htbp]
\centering
\caption{Existing 2D-to-3D evaluation results}
\label{evaluation_results}
\begin{tabular}{lrr} 
\toprule
               Method &  IoU↑ &    CD↓ \\
\midrule
                 Ours & \textbf{0.139} & \textbf{0.038} \\
      Stable Fast 3D\cite{Sf3d2024} & 0.079 & 0.069 \\
Stable Point Aware 3D\cite{SPAR3D_2025} & 0.080 & 0.068 \\
              TripoSR\cite{Triposr2024} & 0.065 & 0.108 \\
\bottomrule
\end{tabular}
\end{table}

The present study compared the proposed method with existing methods for generating 3D meshes from a single 2D image. While front views were used in this study, existing models require isometric view-based inputs for processing. This distinction arises from the limitation of existing models, which struggle to fully reproduce the entire wheel geometry when provided with only front view input.

The \autoref{Comparison existing 2D-to-3D approachs} shows the results of a qualitative comparison. The method used in this study has better mesh consistency than existing techniques and has shown strengths in clearly restoring key structural elements such as spokes and lug holes. On the other hand, the existing methods were observed to have overall quality deterioration, such as the inability to maintain the symmetry and shape of the wheel or the loss of detail in certain parts. 

Stable Fast 3D~\cite{Sf3d2024} and Stable Point Aware 3D~\cite{SPAR3D_2025} maintained the overall shape of the wheel but showed limitations in the expression of details and distorted the restored mesh. TripoSR~\cite{Triposr2024} failed to generate spoke patterns and only left a blocked appearance, resulting in poor overall mesh quality. This is because the methodologies emphasized texture-mapped meshes, while the main objective was not to restore the CAD-level details.

This study used Intersection over Union (IoU) and Chamfer Distance (CD) metrics to evaluate the generated 3D mesh quantitatively. 
IoU calculates the percentage of overlapping areas after converting the generated mesh $M_{\text{pred}}$, and the ground truth (GT) mesh $M_{\text{gt}}$ into volume grids of the exact resolution. The formula is as follows:

\begin{equation}
\text{IoU} = \frac{|M_{\text{pred}} \cap M_{\text{gt}}|}{|M_{\text{pred}} \cup M_{\text{gt}}|}
\end{equation}

The higher the IoU value, the higher the shape similarity with the GT. 

Chamfer Distance (CD) is based on the average shortest distance between uniformly sampled point sets $P$ and $Q$ in the two meshes, and is defined as follows:

\begin{equation}
\text{CD}(P, Q) = \frac{1}{|P|} \sum_{p \in P} \min_{q \in Q} \|p - q\|_2^2 + \frac{1}{|Q|} \sum_{q \in Q} \min_{p \in P} \|q - p\|_2^2
\end{equation}

Here, $P$ and $Q$ are the sets of points uniformly sampled from the predicted mesh and the GT mesh, respectively, and $\| \cdot \|_2$ is the Euclidean distance and $| \cdot |$ is the number of points. The lower the CD, the closer the recovered shape is to the GT. 

As shown in \autoref{evaluation_results}, the method of this study showed the best performance with an IoU of 0.139 and a CD of 0.038. On the other hand, TripoSR showed the lowest IoU of 0.065 and the highest CD of 0.108, indicating a significant deterioration in reconstruction quality. These results suggest that accurate and consistent shape restoration is essential for using 3D models for simulation or prototype production. Realistically reproducing key parts such as rims and spokes is necessary to achieve the desired results. The method used in this study is expected to provide reliable 3D reconstruction results in the mechanical and industrial fields by restoring structural elements more precisely than existing approaches.

\subsection{Design Space Analysis}

 \begin{figure}[!htbp]
    \centering
    \includegraphics[width=1\linewidth]{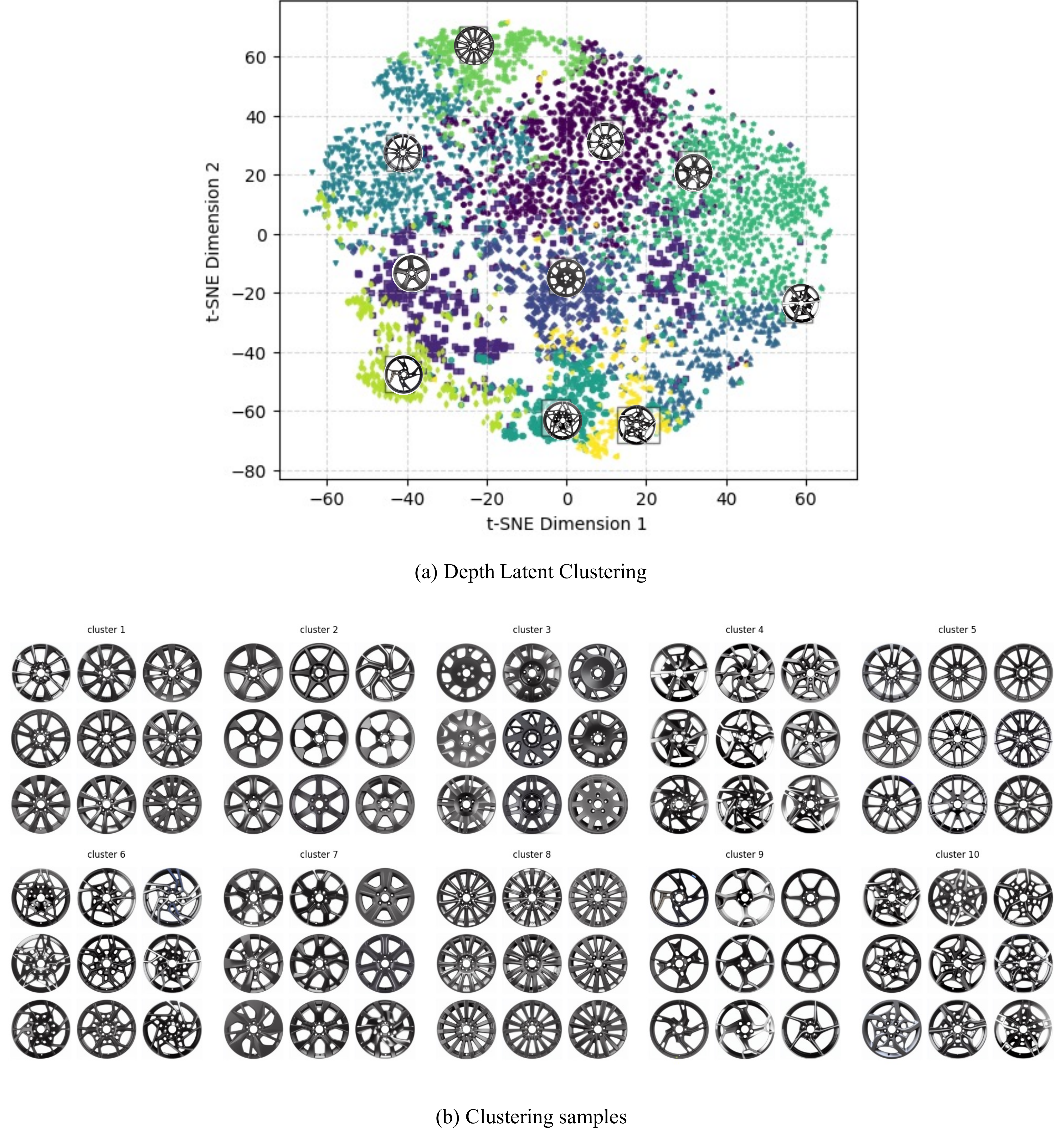}
    \caption{Example of clusters formed in the depth-based 2D embedding}
    \label{depth_cluster_details}
\end{figure}

Efficient design space exploration requires the reduction of generated images (or design samples) to a low-dimensional embedding space followed by balanced sampling. In this study, the Vision Transformer (ViT) \cite{vit2020} was employed to extract high-dimensional features, after which two-dimensional embedding was performed using t-SNE \cite{t-SNE2008}. Latin Hypercube Sampling (LHS) \cite{McKayLHS1979} was subsequently applied to the resulting two-dimensional space, enabling the extraction of a sample set that uniformly covers the entire design space.

The embedding result varies depending on whether RGB is used as the input image or the depth estimated by the depth estimator. RGB is sensitive to surface characteristics such as lighting, color, and material, while Depth emphasizes objects' structural and geometric features. For example, even the same shape can look very different in an RGB image depending on the lighting conditions or color variations. However, similar depth distributions are observed in a depth map, so samples with similar actual shapes are embedded closer together.

In this study, K-means clustering was performed on both RGB-based and depth-based embeddings to compare the impact of these representational differences on clustering outcomes. Subsequently, the quality indices of the resulting clusters were evaluated. This comparative analysis was conducted to determine whether the dimensionality reduction results effectively group similar designs and to identify the more suitable data representation method for design space exploration.

\subsubsection{RGB vs. Depth-based Clustering} 
\begin{table}[!htbp]
    \centering
    \caption{Comparison of clustering performance: rgb vs. depth-based embedding}
    \label{tab:clustering_comparison}
    \begin{tabular}{lcc}
        \toprule
        \textbf{Metric} & \textbf{RGB} & \textbf{Depth} \\
        \midrule
        Silhouette Score (↑) & 0.06 & \textbf{0.10} \\
        Davies-Bouldin Index (↓) & 2.80 & \textbf{2.20} \\
        Calinski-Harabasz Index (↑) & 300.99 & \textbf{471.65} \\
        \bottomrule
    \end{tabular}
\end{table}

Typical clustering performance indicators include the Silhouette Score, Davies-Bouldin Index, and Calinski-Harabasz Index. First, the Silhouette Score has a value between -1 and 1 by taking into account the distance (cohesion) within the cluster to which each data point belongs and the distance (separation) to other clusters, and the closer it is to 1, the more appropriately the data points are clustered. A value close to 0 indicates that the data is located at the cluster's boundary, while a negative value indicates that the data is more likely to be assigned to the wrong cluster. The Davies-Bouldin Index is an average of the distances between the centroids of each cluster and the closest other cluster, and the smaller the value, the better the cohesion within the cluster and the better the separation between different clusters. Finally, the Calinski-Harabasz Index increases as the distance between cluster centers increases, and the intra-cluster variance decreases, and a high value indicates better overall clustering performance.

~\autoref{tab:clustering_comparison} presents the results of K-means clustering applied to RGB and depth embeddings, along with representative cluster quality metrics.

For the Silhouette Score, depth embedding achieved a score of 0.10, approximately 66.7\% higher than RGB, indicating a more precise separation between clusters.
Regarding the Davies-Bouldin Index (DBI), depth embedding recorded 2.20, which is 21.4\% lower than RGB (2.80), suggesting tighter intra-cluster cohesion and greater inter-cluster separation.
The Calinski-Harabasz Index (CH) also favored depth embedding, scoring 471.65, about 56.7\% higher than RGB (300.99). This implies that the structural distribution of designs is well preserved during dimensionality reduction, with similar shapes effectively clustered together.

\autoref{depth_cluster_details}(a) visualizes the RGB images corresponding to the 10 clusters formed in the 2D space after reducing depth embeddings, while (b) shows the cluster centroids along with the nine nearest samples. Depth-based embedding produces more distinct and cohesive clusters than RGB-based clustering, reflecting stronger shape and structural similarity even in the reduced t-SNE space.
This result suggests that depth embeddings provide a more reliable foundation for representative sampling methods such as Latin Hypercube Sampling (LHS), enabling a more uniform and comprehensive design space exploration.

As the design space becomes more complex, sampling tends to become biased toward specific regions, significantly limiting the overall exploration range.
Depth-based embeddings enable more balanced sampling across the entire design space by effectively capturing structural characteristics.

\subsection{Performance Space Analysis}

\begin{figure}[!htbp]
    \centering
    \includegraphics[width=1\linewidth]{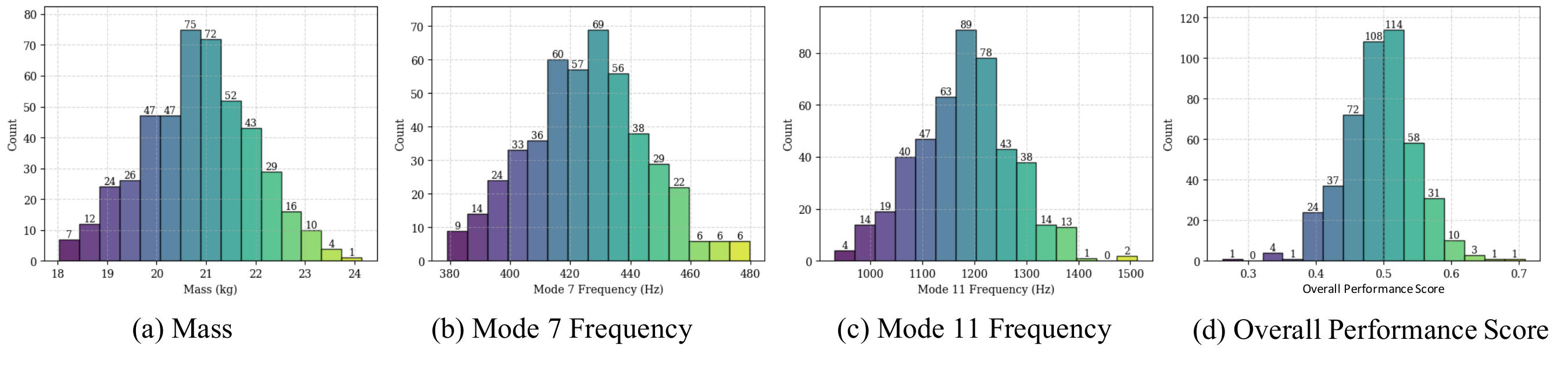}
    \caption{Distribution of overall performance score}
    \label{Performance Index Distribution}
\end{figure}
\begin{figure}[!htbp]
    \centering
    \includegraphics[width=1\linewidth]{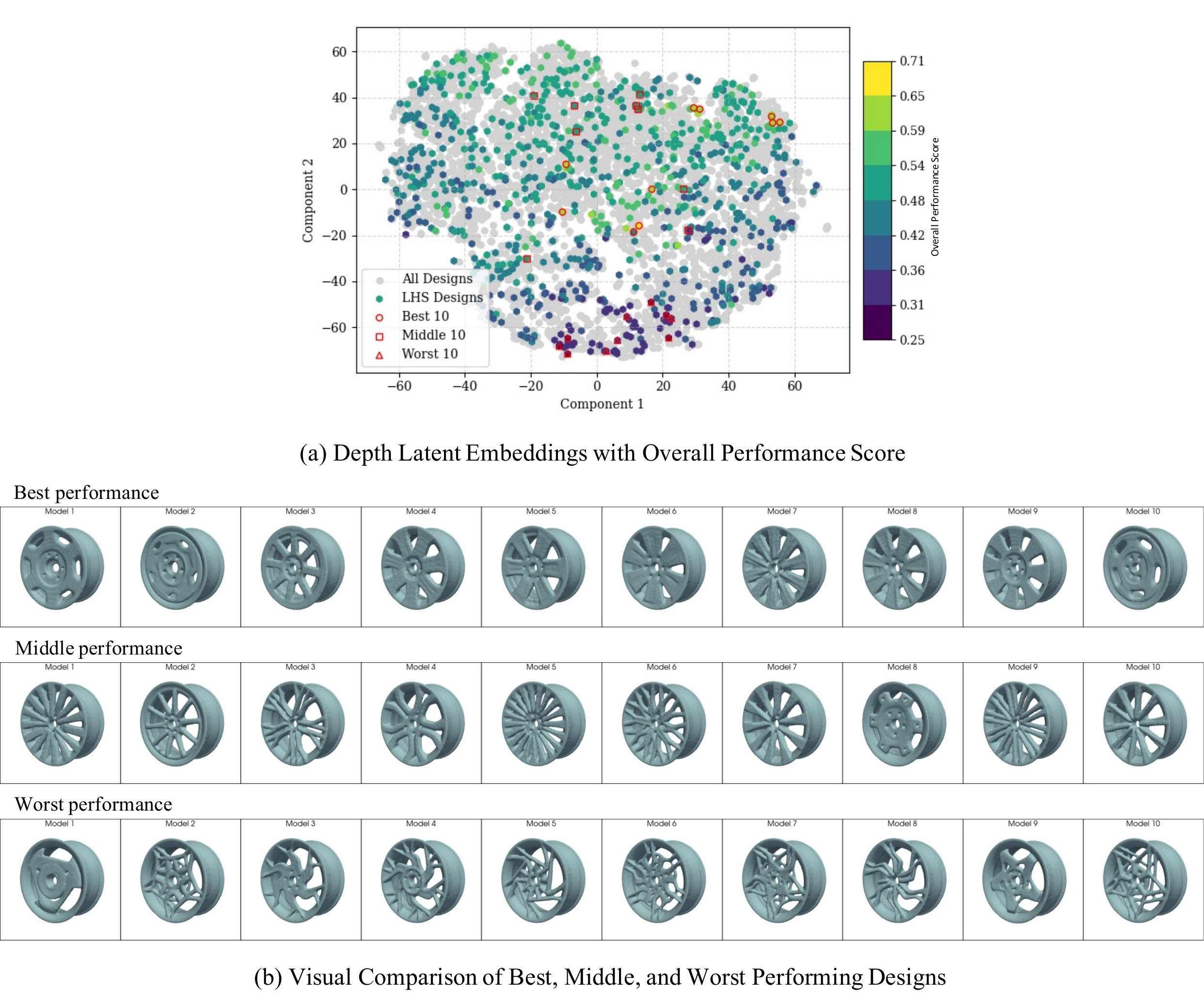}
    \caption{Representative designs with best, middle, and worst score based on the overall performance score}
    \label{Based on the Performance Score}
\end{figure}

Of the 1,000 total design cases, 904 were successfully simulated. The remaining cases were excluded from analysis due to mesh generation or solver errors. The simulation results, including natural frequency and mass, were used to assess the dynamic characteristics of each design quantitatively and will serve as a foundation for future data-driven design exploration and optimization.

\autoref{Performance Index Distribution} visualizes the simulation outcomes of the 904 valid cases. Each subplot shows the distribution of (a) mass, (b) Mode 7 frequency, (c) Mode 11 frequency, and (d) the overall performance score. Both mass and natural frequencies approximate a normal distribution, indicating consistent simulation results across the dataset.

The overall performance score in (d) was computed by individually normalizing the 7th and 11th mode frequencies to values closer to 1 for higher frequencies and the mass to values closer to 1 for smaller masses. These three normalized metrics were then averaged with equal weighting.

\autoref{Based on the Performance Score} visualizes the distribution of designs in the design space based on this integrated performance index. Specifically, ~\ref{Based on the Performance Score}(a) maps the overall performance scores of the designs shown in the same component as ~\ref{depth_cluster_details}(a).

For example, the top 10 designs in three categories, best, middle, and worst, are highlighted. ~\ref{Based on the Performance Score}(b) shows that high- and mid-performance designs are predominantly located in the upper and middle regions of the design space. In contrast, low-performance designs tend to cluster in the lower region. High-performance designs generally feature spoke-based structures commonly seen in off-road wheels, whereas mid-performance designs resemble the spoke layout of luxury sedans. In contrast, low-performance designs include many creatively generated shapes with unconventional, topology-optimized spoke configurations.

\subsection{Design \& Performance Space Diversity}

\begin{figure}[!ht]
    \centering
    \includegraphics[width=1\linewidth]{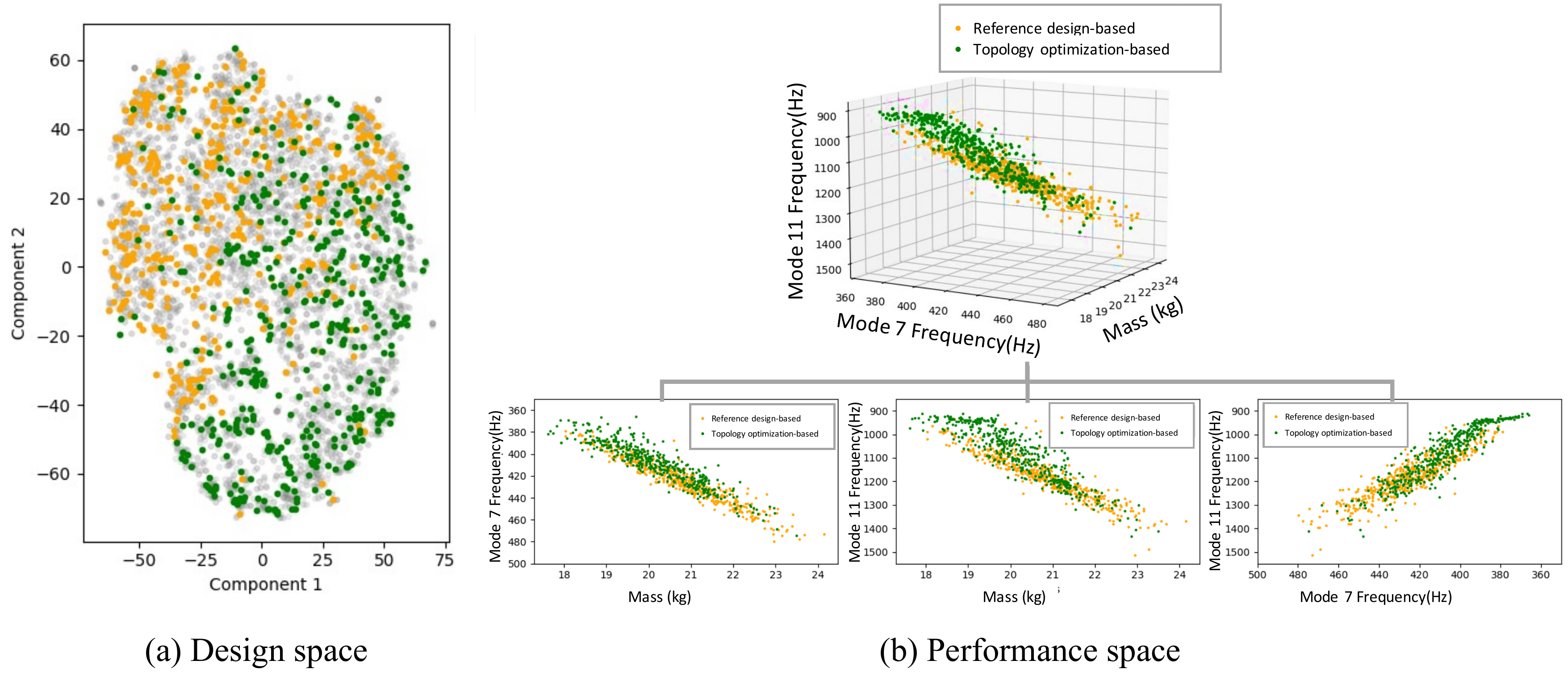}
    \caption{Scatter plot of design \& performance space}
    \label{Scatter Plot of Design Performance Space}
\end{figure}

\begin{table}[!ht]
    \centering
    \caption{Comparison of design space diversity and performance space diversity}
    \label{tab:dsd_psd_comparison}
    \begin{tabular}{lcc}
        \toprule
        \textbf{Metric} & \textbf{Reference Design Space} & \textbf{Overall Design Space} \\
        \midrule
        Design Space Diversity & 6.02 & \textbf{6.45} \\
        Performance Space Diversity & 106.93 & \textbf{124.00} \\
        \bottomrule
    \end{tabular}
\end{table}

To measure how much topology optimization enhances the expansion of design space, a comparative analysis was conducted between the Reference design-based group and the overall design set, including Reference design-based and Topology optimization-based designs. The diversity of the resulting design space was quantified using the Design Space Diversity (DSD) and Performance Space Diversity (PSD) metrics proposed in \cite{DsdPsd2022}. These metrics evaluate the degree of dissimilarity among all design instances by averaging the pairwise distances, as formalized below:

\begin{equation}
s_{\text{div}, i} = \frac{1}{n - 1} \sum_{j \in P} \phi\bigl(v_i,\, v_j\bigr)
\label{diversity}
\end{equation}

Here, $s_{\text{div}, i}$ is the diversity score for sample $i$, and $P$ is a randomly selected subset of $n$ samples. The function $\phi(v_i, v_j)$ computes the Euclidean distance between two vectors. Depending on the context, $v$ represents either a design vector ($y$) or a performance vector ($f$). Specifically, DSD is computed using 1,000 design samples, while PSD is computed using all 904 available performance samples.

A higher PSD value indicates greater variation in performance characteristics, providing designers with a wider range of performance outcomes.

\autoref{tab:dsd_psd_comparison} compares the DSD and PSD values between the Reference design space and the entire design space, including topology-optimized samples. With the introduction of topology optimization, DSD increased from 6.02 to 6.45, and PSD increased from 106.93 to 124.00. This corresponds to a relative increase of 7.25\% in design diversity and 15.97\% in performance diversity. These results suggest that topology optimization finds novel structures beyond the reference design group and enriches the performance space.

\autoref{Scatter Plot of Design Performance Space} illustrates the distribution of the overall design space~\ref{Scatter Plot of Design Performance Space}(a) and performance space~\ref{Scatter Plot of Design Performance Space}(b), with color labels distinguishing between designs generated based on existing reference images and those generated from topology optimization images. The existing reference images refer to wheel designs currently available on the market. The results show that topology optimization enables broader and more effective exploration of design and performance spaces. This expansion allows designers to simultaneously consider multiple design alternatives that meet different performance objectives, enhancing their decision-making flexibility.

\subsection{Qualitative Comparison of Our Results and Previous Research}
\begin{figure}[h!]
    \centering
    \includegraphics[width=1\linewidth]{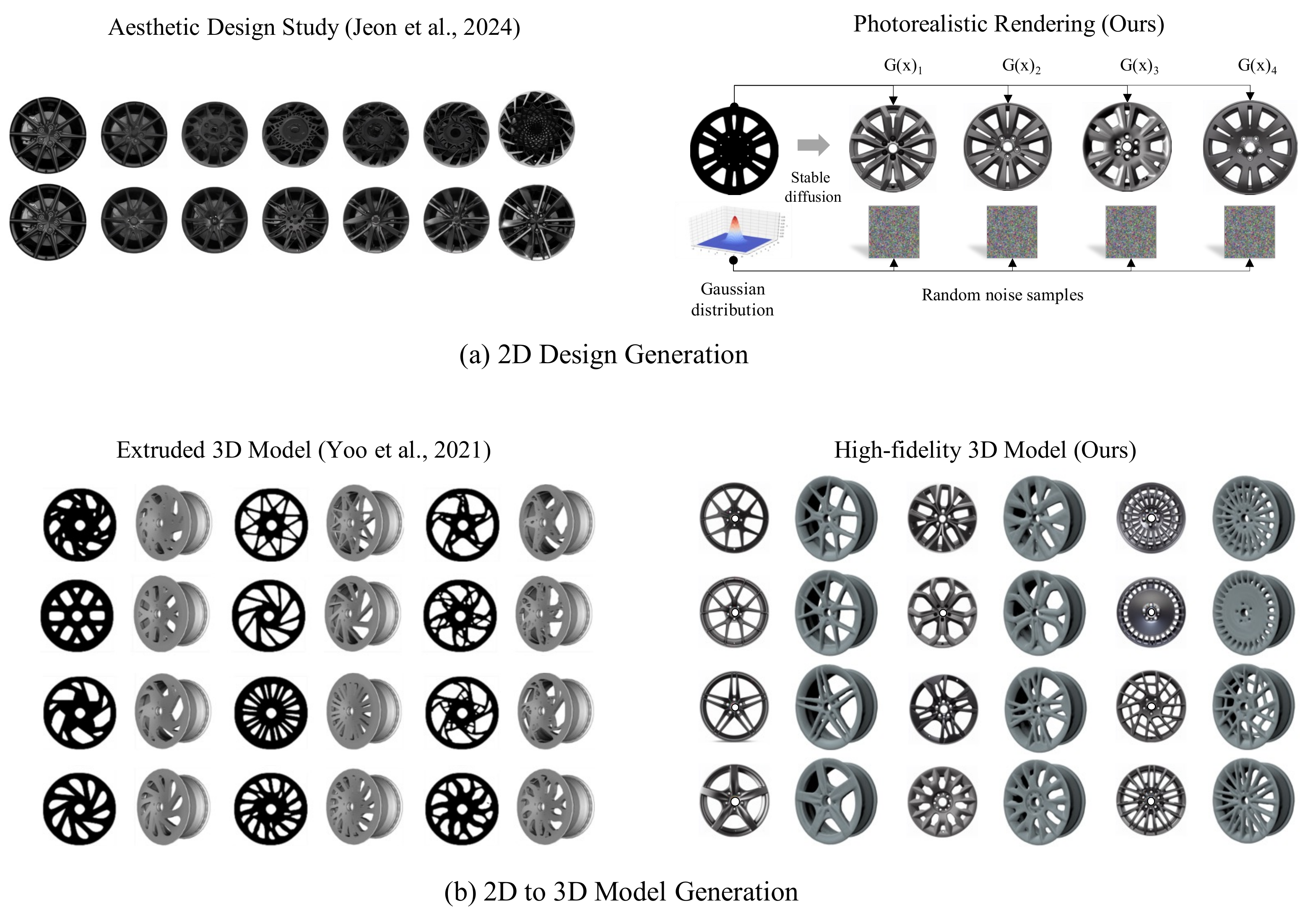}
    \caption{Comparison of wheel design generation and 3D reconstruction results}
    \label{QualitativeComparison}
\end{figure}

~\autoref{QualitativeComparison} presents a visual comparison between previous studies and the results of this study, highlighting qualitative differences in wheel design generation and 3D shape reconstruction. The comparison consists of two parts: (a) 2D design generation and (b) 3D shape reconstruction based on the generated designs.

(a) In the 2D design comparison, our method is evaluated against the Aesthetic Design Study by Jeon et al. (2024)\cite{jeon2024}. Jeon et al. explored the 2D style design space using designers' aesthetic evaluation data, enabling creative idea generation via keyword-based style mixing and comparison tools. However, the geometric accuracy of the output images is limited, as the focus is primarily on visual ideation. In contrast, our study emphasizes geometric realism by rendering realistic wheel designs with various structures, facilitating downstream tasks such as 3D reconstruction and engineering analysis.

(b) In the 3D reconstruction comparison, the 2D–3D conversion method proposed by Yoo et al. (2021)\cite{yoo2021} was contrasted with the depth-based reconstruction approach presented in this study. Yoo et al. (2021)\cite{yoo2021}, developed in prior work, extrudes a 3D model from a simple 2D mask input. While computationally efficient and useful for generating large-scale synthetic datasets, this approach is limited in capturing the complex three-dimensional structures of industrial wheel designs.

To address these limitations, a domain-specific monocular depth prediction model was employed in the current study to generate accurate 2.5D representations from 2D images. These depth maps were subsequently utilized to reconstruct high-resolution 3D mesh models that faithfully reflect fine-grained geometry. The resulting models demonstrate improved realism and structural analysis suitability, enhancing consistency with real simulation data.

\section{Conclusion}

This study presents a comprehensive workflow for constructing a wheel design dataset, encompassing the entire pipeline from 2D rendering-based design generation to 3D shape reconstruction and structural performance analysis. Unlike conventional 2D-to-3D conversion methods, the proposed pipeline enables structurally consistent restoration of key wheel elements. 

Through data augmentation during fine-tuning of a depth prediction foundation model, complex geometries can be reliably predicted even with limited training data, demonstrating the potential for developing powerful domain-specific models.
Depth map-based embeddings were found to enable more balanced sampling than RGB-based representations. Due to their more effective capture of structural features, similar designs remain closely clustered even after dimensionality reduction or clustering. This characteristic is essential for efficient design space exploration and representative sample selection.

From a performance perspective, designs were evaluated using a normalized performance score combining mass and natural frequency. Results indicate that spoke placement and geometry significantly affect performance: high-performance designs resemble off-road wheels, mid-performance ones align with sedan-style wheels, and low-performance samples, often generated via topology optimization, exhibit unique but less efficient structures.

To assess the impact of topology optimization on design space exploration, Design Space Diversity (DSD) and Performance Space Diversity (PSD) metrics were measured. Both metrics increased notably in the topology-augmented set compared to the reference-only set, indicating that topology optimization helps uncover a broader range of structural and performance characteristics, thereby expanding design possibilities.

The workflow integrates the whole process from sketch-level 2D ideation to accurate 3D modeling and simulation-based evaluation. This framework enabled the construction of a large-scale synthetic dataset consisting of diverse modalities, including binary masks, photorealistic renderings, 3D meshes, and performances. 

Such datasets serve as a foundational infrastructure for AI-based design research, addressing prior limitations related to data scarcity and format inconsistency. The automated pipeline produces high-quality data at scale and is suitable for real-world engineering tasks, opening new opportunities. From an industrial standpoint, this workflow can be extended to design problems involving complex geometries beyond wheels. Its integration into a tailored pipeline can streamline 3D dataset generation and significantly reduce development time and cost.

Future work will expand this framework to support multi-view input rather than a single image, thereby reconstructing more complex shapes. Multi-view methods are expected to capture detailed structures and curved surfaces better, facilitating richer 3D and performance data generation across broader product domains.

\section{Licensing, Attribution, and Accessibility}
\label{sec:licensing}

The \textbf{DeepWheel dataset} is a publicly available resource designed to facilitate research in 3D wheel design, reconstruction, and structural performance analysis. It comprises multiple data types generated through an integrated pipeline involving image-based rendering, depth estimation, 3D reconstruction, and simulation. An overview of the dataset components is provided in Table~\ref{tab:deepwheel_dataset}.

\begin{table}[!ht]
    \centering
    \caption{Overview of DeepWheel dataset components}
    \label{tab:deepwheel_dataset}
    \renewcommand{\arraystretch}{1.3}
    \begin{tabularx}{\linewidth}{@{}l l X@{}}
        \toprule
        \textbf{Category} & \textbf{Path (base: \texttt{./deepwheel\_1.0/})} & \textbf{Description} \\
        \midrule
        Rendered Images &
        1\_rendered\_images &
        6,249 RGB images generated using Stable Diffusion. \\
        
        Predicted Depth Maps &
        2\_predicted\_depth\_maps &
        6,249 single-channel grayscale depth maps aligned with the rendered images. \\
        
        3D Reconstruction Meshes &
        3\_3D\_recon\_meshes &
        904 STL mesh files reconstructed from the predicted depth maps. \\
        
        3D CAD Models &
        4\_3D\_cad\_models &
        904 STEP files converted from the reconstructed meshes for CAE analysis. \\
        
        Simulation Results &
        deepwheel\_sim\_results.csv &
        Modal analysis results, including structural mass and natural frequencies. \\
        \bottomrule
    \end{tabularx}
    \vspace{0.5em}
\end{table}

\paragraph{Generation Process}
All 2D renderings were created using the Stable Diffusion v1.4 model~\cite{CompVisStableDiffusionV14}, released under the CreativeML Open RAIL-M license. The model was adapted for wheel rendering by conditioning it on binary masks derived from reference and topology-optimized designs. A fine-tuned depth estimation model was then applied to generate corresponding 2.5D depth maps. These depth maps were converted into 3D surface meshes and subsequently transformed into CAD-compatible STEP models. Structural properties were evaluated through modal analysis, and the results were consolidated into a CSV file.

\paragraph{Hosting and Compatibility}
The dataset is publicly hosted on Google Drive and accessible via the following link: \href{https://www.smartdesignlab.org/datasets}{\texttt{smartdesignlab.org/datasets}}. All 3D data files are compatible with standard commercial and open-source CAD/CAE platforms, as well as Python-based simulation environments.

\paragraph{License}
The DeepWheel dataset is distributed under the Creative Commons Attribution-NonCommercial 4.0 International (CC BY-NC 4.0) license. This license permits use, modification, and redistribution for non-commercial purposes, provided proper attribution is given. Any commercial use of the dataset or derivative models is strictly prohibited.

\paragraph{Availability and Maintenance}
The dataset will remain publicly accessible for at least ten years following its release. If any updates, revisions, or corrections are made, they will be announced with appropriate version control and a revised access link. A \texttt{README} file is included to assist users in navigating the dataset.

\section{Acknowledgments}

This work was supported by the Ministry of Science and ICT of Korea grant (No. 2022-0-00969, No. 2022-0-00986, and No. GTL24031-000) and the Ministry of Trade, Industry \& Energy grant (RS-2024-00410810).

\bibliographystyle{IEEEtran}
\bibliography{references}  

\begin{thebibliography}{10}
\providecommand{\url}[1]{#1}
\csname url@samestyle\endcsname
\providecommand{\newblock}{\relax}
\providecommand{\bibinfo}[2]{#2}
\providecommand{\BIBentrySTDinterwordspacing}{\spaceskip=0pt\relax}
\providecommand{\BIBentryALTinterwordstretchfactor}{4}
\providecommand{\BIBentryALTinterwordspacing}{\spaceskip=\fontdimen2\font plus
\BIBentryALTinterwordstretchfactor\fontdimen3\font minus \fontdimen4\font\relax}
\providecommand{\BIBforeignlanguage}[2]{{%
\expandafter\ifx\csname l@#1\endcsname\relax
\typeout{** WARNING: IEEEtran.bst: No hyphenation pattern has been}%
\typeout{** loaded for the language `#1'. Using the pattern for}%
\typeout{** the default language instead.}%
\else
\language=\csname l@#1\endcsname
\fi
#2}}
\providecommand{\BIBdecl}{\relax}
\BIBdecl

\bibitem{damen2024exploring}
N.~B. Damen, V.~Seo, and Y.~Wang, ``Exploring opportunities for adopting generative ai in automotive conceptual design,'' in \emph{International Design Engineering Technical Conferences and Computers and Information in Engineering Conference}, vol. 88346.\hskip 1em plus 0.5em minus 0.4em\relax American Society of Mechanical Engineers, 2024, p. V02AT02A051.

\bibitem{feldinger2017automotive}
U.~E. Feldinger, S.~Kleemann, T.~Vietor \emph{et~al.}, ``Automotive styling: Supporting engineering-styling convergence through surface-centric knowledge based engineering,'' in \emph{DS 87-4 Proceedings of the 21st International Conference on Engineering Design (ICED 17) Vol 4: Design Methods and Tools, Vancouver, Canada, 21-25.08. 2017}, 2017, pp. 139--148.

\bibitem{yoo2021}
S.~Yoo, S.~Lee, S.~Kim, K.~H. Hwang, J.~H. Park, and N.~Kang, ``Integrating deep learning into cad/cae system: Generative design and evaluation of 3d conceptual wheel,'' \emph{Structural and Multidisciplinary Optimization}, vol.~64, no.~4, pp. 2725--2747, 2021.

\bibitem{song2023Surrogate}
B.~Song, C.~Yuan, F.~Permenter, N.~Arechiga, and F.~Ahmed, ``Surrogate modeling of car drag coefficient with depth and normal renderings,'' in \emph{International Design Engineering Technical Conferences and Computers and Information in Engineering Conference}, vol. 87301.\hskip 1em plus 0.5em minus 0.4em\relax American Society of Mechanical Engineers, 8 2023, p. V03AT03A029.

\bibitem{shin2023wheel}
S.~Shin, A.-h. Jin, S.~Yoo, S.~Lee, C.~Kim, S.~Heo, and N.~Kang, ``Wheel impact test by deep learning: prediction of location and magnitude of maximum stress,'' \emph{Structural and Multidisciplinary Optimization}, vol.~66, no.~1, p.~24, 2023.

\bibitem{shin2024uda}
S.~Shin and N.~Kang, ``Weighted unsupervised domain adaptation considering geometry features and engineering performance of 3d design data,'' \emph{Expert Systems with Applications}, vol. 256, p. 124928, 2024.

\bibitem{kim2025physics}
J.~Kim, J.~Park, N.~Kim, Y.~Yu, K.~Chang, C.-S. Woo, S.~Yang, and N.~Kang, ``Physics-constrained graph neural networks for spatio-temporal prediction of drop impact on oled display panels,'' \emph{Expert Systems with Applications}, p. 126907, 2025.

\bibitem{park2024bmo}
J.~Park and N.~Kang, ``Bmo-gnn: Bayesian mesh optimization for graph neural networks to enhance engineering performance prediction,'' \emph{Journal of Computational Design and Engineering}, vol.~11, no.~6, pp. 260--271, 2024.

\bibitem{song2024multi}
B.~Song, R.~Zhou, and F.~Ahmed, ``Multi-modal machine learning in engineering design: A review and future directions,'' \emph{Journal of Computing and Information Science in Engineering}, vol.~24, no.~1, p. 010801, 2024.

\bibitem{rad2024datasets}
M.~A. Rad, T.~Hajali, J.~M. Bonde, M.~Panarotto, K.~Wärmefjord, J.~Malmqvist, and O.~Isaksson, ``Datasets in design research: needs and challenges and the role of ai and gpt in filling the gaps,'' in \emph{Proceedings of the Design Society}, vol.~4, 2024, pp. 1919--1928.

\bibitem{elrefaie2025drivaernet++}
M.~Elrefaie, F.~Morar, A.~Dai, and F.~Ahmed, ``Drivaernet++: A large-scale multimodal car dataset with computational fluid dynamics simulations and deep learning benchmarks,'' \emph{Advances in Neural Information Processing Systems}, vol.~37, pp. 499--536, 2025.

\bibitem{li2022predictive}
X.~Li, C.~Xie, and Z.~Sha, ``A predictive and generative design approach for three-dimensional mesh shapes using target-embedding variational autoencoder,'' \emph{Journal of Mechanical Design}, vol. 144, no.~11, p. 114501, 2022.

\bibitem{Get3d2022}
J.~Gao, T.~Shen, Z.~Wang, W.~Chen, K.~Yin, D.~Li, and S.~Fidler, ``Get3d: A generative model of high quality 3d textured shapes learned from images,'' \emph{Advances in Neural Information Processing Systems}, vol.~35, pp. 31\,841--31\,854, 2022.

\bibitem{Magic3d2023}
C.~H. Lin, J.~Gao, L.~Tang, T.~Takikawa, X.~Zeng, X.~Huang, and T.~Y. Lin, ``Magic3d: High-resolution text-to-3d content creation,'' in \emph{Proceedings of the IEEE/CVF Conference on Computer Vision and Pattern Recognition}, 2023, pp. 300--309.

\bibitem{PointE2022}
A.~Nichol, H.~Jun, P.~Dhariwal, P.~Mishkin, and M.~Chen, ``Point-e: A system for generating 3d point clouds from complex prompts,'' \emph{arXiv preprint arXiv:2212.0}, 2022.

\bibitem{Zero-1-to-3_2023}
R.~Liu, R.~Wu, B.~Van~Hoorick, P.~Tokmakov, S.~Zakharov, and C.~Vondrick, ``Zero-1-to-3: Zero-shot one image to 3d object,'' in \emph{Proceedings of the IEEE/CVF International Conference on Computer Vision}, 2023, pp. 9298--9309.

\bibitem{One-2-3-45_2023}
M.~Liu, C.~Xu, H.~Jin, L.~Chen, T.~M. Varma, Z.~Xu, and H.~Su, ``One-2-3-45: Any single image to 3d mesh in 45 seconds without per-shape optimization,'' \emph{Advances in Neural Information Processing Systems}, vol.~36, pp. 22\,226--22\,246, 2023.

\bibitem{Triposr2024}
D.~Tochilkin, D.~Pankratz, Z.~Liu, Z.~Huang, A.~Letts, Y.~Li, and Y.~P. Cao, ``Triposr: Fast 3d object reconstruction from a single image,'' \emph{arXiv preprint arXiv:2403.02151}, 2024.

\bibitem{Sf3d2024}
M.~Boss, Z.~Huang, A.~Vasishta, and V.~Jampani, ``Sf3d: Stable fast 3d mesh reconstruction with uv-unwrapping and illumination disentanglement,'' \emph{arXiv preprint arXiv:2408.00653}, 2024.

\bibitem{deitke2023objaverse}
M.~Deitke, D.~Schwenk, J.~Salvador, L.~Weihs, O.~Michel, E.~VanderBilt, and A.~Farhadi, ``Objaverse: A universe of annotated 3d objects,'' in \emph{Proceedings of the IEEE/CVF Conference on Computer Vision and Pattern Recognition}, 2023, pp. 13\,142--13\,153.

\bibitem{chang2015shapenet}
A.~X. Chang, T.~Funkhouser, L.~Guibas, P.~Hanrahan, Q.~Huang, Z.~Li, and F.~Yu, ``Shapenet: An information-rich 3d model repository,'' \emph{arXiv preprint arXiv:1512.03012}, 2015.

\bibitem{rosset2023}
N.~Rosset, G.~Cordonnier, R.~Duvigneau, and A.~Bousseau, ``Interactive design of 2d car profiles with aerodynamic feedback,'' in \emph{Computer Graphics Forum}, vol.~42, 5 2023, pp. 427--437.

\bibitem{willis2021fusion}
K.~D. Willis, Y.~Pu, J.~Luo, H.~Chu, T.~Du, J.~G. Lambourne, A.~Solar-Lezama, and W.~Matusik, ``Fusion 360 gallery: A dataset and environment for programmatic cad construction from human design sequences,'' \emph{ACM Transactions on Graphics (TOG)}, vol.~40, no.~4, 2021.

\bibitem{bagazinski2023ship}
N.~J. Bagazinski and F.~Ahmed, ``Ship-d: Ship hull dataset for design optimization using machine learning,'' in \emph{International Design Engineering Technical Conferences and Computers and Information in Engineering Conference}, vol. 87301.\hskip 1em plus 0.5em minus 0.4em\relax American Society of Mechanical Engineers, 8 2023, p. V03AT03A028.

\bibitem{wollstadt2022carhoods10k}
P.~Wollstadt, M.~Bujny, S.~Ramnath, J.~J. Shah, D.~Detwiler, and S.~Menzel, ``Carhoods10k: An industry-grade dataset for representation learning and design optimization in engineering applications,'' \emph{IEEE Transactions on Evolutionary Computation}, vol.~26, no.~6, pp. 1221--1235, 2022.

\bibitem{whalen2021simjeb}
E.~Whalen, A.~Beyene, and C.~Mueller, ``Simjeb: Simulated jet engine bracket dataset,'' \emph{Computer Graphics Forum}, vol.~40, no.~5, pp. 9--17, 8 2021.

\bibitem{hong2025deepjeb}
S.~Hong, Y.~Kwon, D.~Shin, J.~Park, and N.~Kang, ``Deepjeb: 3d deep learning-based synthetic jet engine bracket dataset,'' \emph{Journal of Mechanical Design}, vol. 147, no.~4, 2025.

\bibitem{oh2019}
S.~Oh, Y.~Jung, S.~Kim, I.~Lee, and N.~Kang, ``Deep generative design: Integration of topology optimization and generative models,'' \emph{Journal of Mechanical Design}, vol. 141, no.~11, p. 111405, 2019.

\bibitem{jang2022}
S.~Jang, S.~Yoo, and N.~Kang, ``Generative design by reinforcement learning: Enhancing the diversity of topology optimization designs,'' \emph{Computer-Aided Design}, vol. 146, p. 103225, 2022.

\bibitem{jeon2024}
Y.~Jeon, M.~K. Hong, Y.~Y. Chen, K.~Murakami, J.~Q. Li, X.~A. Chen, and M.~Klenk, ``Weaving ml with human aesthetic assessments to augment design space exploration: An automotive wheel design case study,'' in \emph{Extended Abstracts of the CHI Conference on Human Factors in Computing Systems}, 5 2024, pp. 1--10.

\bibitem{wang2024}
Y.~Wang, N.~B. Damen, T.~Gale, V.~Seo, and H.~Shayani, ``Inspired by ai? a novel generative ai system to assist conceptual automotive design,'' in \emph{International Design Engineering Technical Conferences and Computers and Information in Engineering Conference}, vol. 88407.\hskip 1em plus 0.5em minus 0.4em\relax American Society of Mechanical Engineers, 8 2024, p. V006T06A030.

\bibitem{li2024}
K.~Y. Li, C.~K. Huang, Q.~W. Chen, H.~C. Zhang, and T.~T. Tang, ``Leveraging generative ai and cad automation for efficient automotive wheel design with limited data,'' 2024, manuscript in preparation.

\bibitem{Instantmesh2024}
J.~Xu, W.~Cheng, Y.~Gao, X.~Wang, S.~Gao, and Y.~Shan, ``Instantmesh: Efficient 3d mesh generation from a single image with sparse-view large reconstruction models,'' \emph{arXiv preprint arXiv:2404.07191}, 2024.

\bibitem{Meshformer_2025}
M.~Liu, C.~Zeng, X.~Wei, R.~Shi, L.~Chen, C.~Xu, and H.~Su, ``Meshformer: High-quality mesh generation with 3d-guided reconstruction model,'' \emph{Advances in Neural Information Processing Systems}, vol.~37, pp. 59\,314--59\,341, 2025.

\bibitem{M3D2024}
L.~Zhang, P.~Shrestha, Y.~Zhou, C.~Xie, and I.~Kitahara, ``M3d: Dual-stream selective state spaces and depth-driven framework for high-fidelity single-view 3d reconstruction,'' \emph{arXiv preprint arXiv:2411.12635}, 2024.

\bibitem{Kim_mass_2024}
J.~Kim, Y.~Kwon, and N.~Kang, ``Deep generative design for mass production,'' \emph{arXiv preprint arXiv:2403.12098}, 2024.

\bibitem{MiDaS2020}
R.~Ranftl, K.~Lasinger, D.~Hafner, K.~Schindler, and V.~Koltun, ``Towards robust monocular depth estimation: Mixing datasets for zero-shot cross-dataset transfer,'' \emph{IEEE Transactions on Pattern Analysis and Machine Intelligence}, vol.~44, no.~3, pp. 1623--1637, 2020.

\bibitem{DepthAnything2024}
L.~Yang, B.~Kang, Z.~Huang, X.~Xu, J.~Feng, and H.~Zhao, ``Depth anything: Unleashing the power of large-scale unlabeled data,'' in \emph{Proceedings of the IEEE/CVF Conference on Computer Vision and Pattern Recognition}, 2024, pp. 10\,371--10\,381.

\bibitem{Marigold2024}
B.~Ke, A.~Obukhov, S.~Huang, N.~Metzger, R.~C. Daudt, and K.~Schindler, ``Repurposing diffusion-based image generators for monocular depth estimation,'' in \emph{Proceedings of the IEEE/CVF Conference on Computer Vision and Pattern Recognition}, 2024, pp. 9492--9502.

\bibitem{Rombach2022}
R.~Rombach, A.~Blattmann, D.~Lorenz, P.~Esser, and B.~Ommer, ``High-resolution image synthesis with latent diffusion models,'' in \emph{Proceedings of the IEEE/CVF Conference on Computer Vision and Pattern Recognition}, 2022, pp. 10\,684--10\,695.

\bibitem{ho2022classifierfree}
J.~Ho and T.~Salimans, ``Classifier-free diffusion guidance,'' \emph{arXiv preprint arXiv:2207.12598}, 2022.

\bibitem{Salmon2024}
D.~A. Salmon, S.~Zhang, R.~Hu, Z.~Yan, S.~Savarese, and J.~Ho, ``Latent consistency models: Synthesizing high-resolution images with few-step inference,'' \emph{arXiv preprint arXiv:2305.13884}, 2024.

\bibitem{wheel_size}
\BIBentryALTinterwordspacing
{Wheel-Size.com}, ``{Wheel-Size.com: The online wheel and tire fitment guide},'' 2025, accessed April 9, 2025. [Online]. Available: \url{https://www.wheel-size.com/}
\BIBentrySTDinterwordspacing

\bibitem{rhino3d}
{Robert McNeel \& Associates}, ``{Rhinoceros 3D – Design, Model, Present, Analyze, Realize},'' \url{https://www.rhino3d.com/}, 2025, accessed April 9, 2025.

\bibitem{simlab}
{Altair Engineering Inc.}, ``{Altair SimLab – Multiphysics Workflows for Simulation-Driven Design},'' \url{https://altair.com/simlab/}, 2025, accessed April 9, 2025.

\bibitem{optistruct}
{Altair Engineering, Inc.}, ``{OptiStruct – Structural Analysis and Optimization Solver},'' \url{https://altair.com/optistruct}, 2025, accessed April 9, 2025.

\bibitem{eigen2014depth}
D.~Eigen, C.~Puhrsch, and R.~Fergus, ``Depth map prediction from a single image using a multi-scale deep network,'' in \emph{Advances in Neural Information Processing Systems (NeurIPS)}, 2014, pp. 2366--2374.

\bibitem{SPAR3D_2025}
Z.~Huang, M.~Boss, A.~Vasishta, J.~M. Rehg, and V.~Jampani, ``Spar3d: Stable point-aware reconstruction of 3d objects from single images,'' \emph{arXiv preprint arXiv:2501.04689}, 2025.

\bibitem{vit2020}
A.~Dosovitskiy, L.~Beyer, A.~Kolesnikov, D.~Weissenborn, X.~Zhai, T.~Unterthiner, and N.~Houlsby, ``An image is worth 16x16 words: Transformers for image recognition at scale,'' \emph{arXiv preprint arXiv:2010.11929}, 2020.

\bibitem{t-SNE2008}
L.~Van~der Maaten and G.~Hinton, ``Visualizing data using t-sne,'' \emph{Journal of Machine Learning Research}, vol.~9, no.~11, 2008.

\bibitem{McKayLHS1979}
M.~D. McKay, R.~J. Beckman, and W.~J. Conover, ``A comparison of three methods for selecting values of input variables in the analysis of output from a computer code,'' \emph{Technometrics}, vol.~42, no.~1, pp. 55--61, 2000.

\bibitem{DsdPsd2022}
L.~Regenwetter and F.~Ahmed, ``Design target achievement index: A differentiable metric to enhance deep generative models in multi-objective inverse design,'' in \emph{International Design Engineering Technical Conferences and Computers and Information in Engineering Conference}, vol. 86236.\hskip 1em plus 0.5em minus 0.4em\relax American Society of Mechanical Engineers, 8 2022, p. V03BT03A046.

\bibitem{CompVisStableDiffusionV14}
CompVis, ``Stable diffusion v1.4,'' \url{https://huggingface.co/CompVis/stable-diffusion-v1-4}, 2022, hugging Face platform. Accessed on April 10, 2025.

\end{thebibliography}


\end{document}